\documentclass[lettersize,journal]{IEEEtran}

\usepackage{cite}
\usepackage{amsmath,amssymb,amsfonts}
\usepackage{algorithmic}
\usepackage{graphicx}
\usepackage{textcomp}
\usepackage{xcolor}
\usepackage{wrapfig}
\usepackage{tikz}
\usepackage{array}
\usepackage[caption=false,font=normalsize,labelfont=sf,textfont=sf]{subfig}
\usepackage{stfloats}
\usepackage{url}
\usepackage{verbatim}
\usepackage{balance}
\def\BibTeX{{\rm B\kern-.05em{\sc i\kern-.025em b}\kern-.08em
    T\kern-.1667em\lower.7ex\hbox{E}\kern-.125emX}}
\usepackage{orcidlink}
\usepackage{xcolor}
\newcommand{\nic}[1]{#1}
\DeclareMathOperator{\atan}{atan}

\def\BibTeX{{\rm B\kern-.05em{\sc i\kern-.025em b}\kern-.08em
    T\kern-.1667em\lower.7ex\hbox{E}\kern-.125emX}}
\begin{document}

\title{RAGE: A Tightly Coupled Radar-Aided Grip Estimator \\ For Autonomous Race Cars}

\author{
    \IEEEauthorblockN{Davide Malvezzi\orcidlink{0009-0003-7603-2025},}
    \IEEEauthorblockN{Nicola Musiu\orcidlink{0009-0001-6832-0997},}
    \IEEEauthorblockN{Eugenio Mascaro,}
    \IEEEauthorblockN{Francesco Iacovacci}
    \IEEEauthorblockN{and Marko Bertogna\orcidlink{0000-0003-2115-4853}}
}

\maketitle
\begin{abstract}
\nic{Real-time estimation of vehicle–tire–road interaction and grip level is critical for allowing autonomous race cars to safely and effectively operate at their physical limits. Traditional approaches to measure tire grip often depend on costly, specialized sensors that require custom installation, limiting scalability and deployment. In this work, we introduce a novel real-time Radar-Aided Grip Estimator (RAGE) that jointly estimates the vehicle velocity vector, tire slip angles, lateral tire forces, and tire model parameters, using only standard sensors, such as IMUs and RADARs, which are commonly available on most of modern autonomous platforms. We validate our approach through both high-fidelity simulations and real-world experiments conducted on the EAV-24 autonomous race car, demonstrating the effectiveness of our method in estimating the vehicle lateral dynamics.}

\end{abstract}


\section{Introduction}

Lateral velocity, slip angles, and lateral tire forces are fundamental quantities in vehicle lateral dynamics, particularly under high lateral acceleration and when tires are operating in the non-linear region. Accurate estimation of these quantities is critical for ensuring vehicle stability \cite{LOP_UKF}, optimizing performance, and enabling advanced control strategies \cite{RAJI_PLANNING}.

Traditionally, lateral velocity is estimated using high-precision and often expensive sensors, such as optical systems. While highly accurate, these systems are not practical for widespread deployment, particularly on production vehicles. Recent research has explored the use of lower-cost alternatives, such as LiDAR and camera-based approaches, to estimate vehicle motion, showing encouraging results.

RADAR sensors are gaining increasing attention for ego-motion estimation, as they offer a compelling trade-off between cost and performance. This makes them attractive not only for high-performance motor-sport applications, but also for future deployment in production vehicles. Using RADAR for lateral velocity estimation can enable robust and scalable solutions capable of detecting and managing critical edge cases, such as loss of traction, in everyday driving scenarios.

\nic{In this work, we introduce RAGE (Radar-Aided Grip Estimator), a novel estimation framework that jointly estimates the vehicle velocity vector, tire slip angles, lateral tire forces, and tire model parameters.}

RAGE is able to accurately estimate the vehicle’s longitudinal and lateral velocity components using only measurements from standard on-board sensors: inertial measurements from an IMU and Doppler velocity data from multiple RADARs. \nic{Since these sensors are commonly available on modern autonomous platforms and no specialized hardware is required, the proposed approach is both cost-effective and broadly applicable.}
\nic{In the proposed framework, tire grip represent the friction level ($\mu$) at the tire–road interface and it is inferred from the estimated tire model parameters which reflects the maximum transferable lateral force in a model-dependent manner.
In particular,}
RAGE employs the empirical Pacejka tire model \cite{PACEJKA} to capture the non-linear relationship between slip and lateral force. Unlike traditional methods with fixed parameters, RAGE estimates tire model coefficients online, enabling real-time adaptation to changing surface conditions, thermal effects, and tire pressure variations.

\begin{figure}[t]
\centering
\includegraphics[width=\linewidth,trim={10cm 2cm 0 11cm},clip]{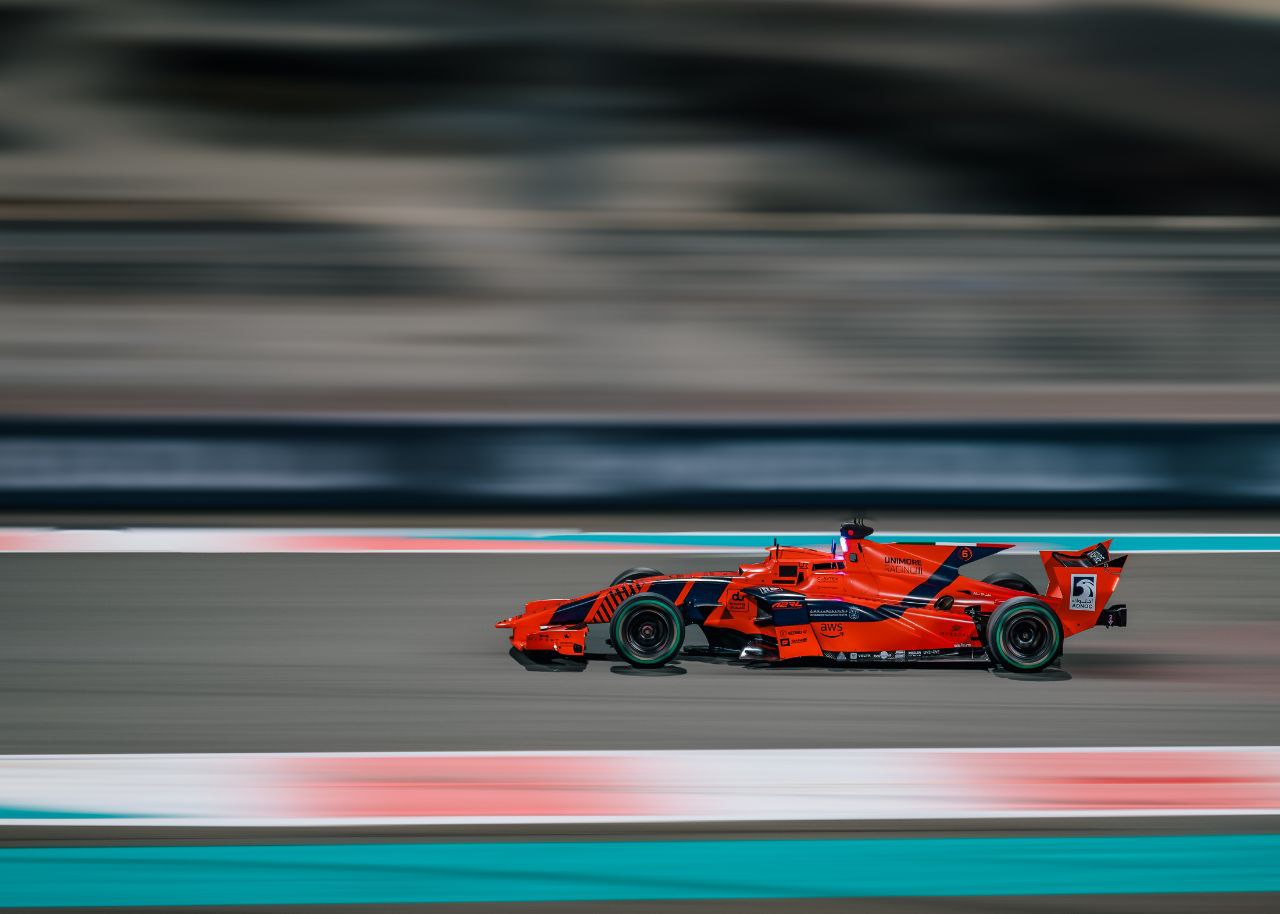}
\caption{\nic{Unimore Racing’s Dallara EAV-24 during the 2024 Abu Dhabi Autonomous Racing League at Yas Marina F1 circuit. Photo by Ian Bundey.}}
\label{fig:car}
\end{figure}

Lateral tire forces are highly dependent on slip angles, which are themselves functions of vehicle speed and steering input. Due to the strong interdependence among these variables, we avoid a sequential estimation, where velocities are first estimated and then used to infer tire model parameters. Instead, we adopt a tightly coupled formulation that simultaneously estimates both the states and the parameters.
In this way, RAGE can capture the evolving dynamics of the vehicle-tire-road interaction more accurately, reducing the uncertainty of the overall estimation. The joint estimation is formulated and solved as a Moving Horizon Estimation (MHE) problem, which optimizes over a sliding time window to provide states and parameters estimates that are consistent with the current sensor data and plausible over time.

The proposed method is first validated in a simulated environment based on a multi-body model \cite{ER_AUTOPILOT}  with highly accurate vehicle dynamics, providing a controlled setting to assess its performance under high-speed conditions. Following simulation, the estimator is further evaluated based on real-world data collected with the EAV-24 autonomous race car (see Figure \ref{fig:car}) during the Abu Dhabi Autonomous Racing League (A2RL\footnote{www.a2rl.io}), where we competed as the Unimore Racing Team\footnote{www.unimoreracing.it}. The dataset includes high-speed driving scenarios under diverse operating conditions, with velocities reaching up to 70 m/s and lateral accelerations up to 28 $\text{m/s}^2$. This provides a comprehensive benchmark for evaluating the proposed estimation framework in a range of dynamic regimes.

The structure of this paper is as follows. Section \ref{sec:related_works} reviews the most relevant related works. Section \ref{sec:overview} presents an overview of the proposed RAGE estimator. The core contributions are detailed in Sections \ref{sec:model} through \ref{sec:lateral}, covering 
\nic{ vehicle model representations,}
IMU bias initialization, Doppler velocity integration, and tire model coefficient estimation. Section \ref{sec:simulation} evaluates the performance of the estimator in simulation, while Section \ref{sec:results} presents experimental results on real-world data. Finally, Section \ref{sec:conclusion} concludes the paper and outlines the directions for future research.
\section{Related Work} \label{sec:related_works}
Direct measurement of lateral velocity and slip angles typically requires expensive instrumentation, including optical sensors or high-precision inertial navigation systems. Consequently, model-based estimation techniques that leverage readily available sensor data, such as yaw rate, accelerations, wheel speeds, and steering angle, have emerged as practical and effective alternatives. These model-based observers often employ kinematic approaches, as in \cite{KINEMATIC_MODEL}, or hybrid methods that combine kinematic and dynamic models \cite{DYNAMIC_KINEMATIC_MODEL}, which offer improved accuracy and adaptability.

Other approaches aim to estimate dynamic lateral forces by incorporating tire models, like Dugoff model \cite{DYNAMIC_KINEMATIC_MODEL2} or Pacejka model \cite{MODEL_BASED}, though these typically rely on precomputed parameters. Other frameworks, like T.R.I.C.K 2.0 \cite{TRICK2}, model vehicle dynamics in great detail. However, these methods require a comprehensive suite of in-vehicle sensors, limiting their practical deployment due to high implementation complexity.

In some cases, neural networks-based methods have outperformed traditional state-of-the-art methods in racing scenarios \cite{NN1}, demonstrating robustness even under varying tire wear and grip conditions in road vehicles \cite{NN2}. However, their effectiveness relies on extensive training data and they often struggle to adapt to unseen system conditions \cite{NN3}, limiting their generalization across different vehicles and environments.

Other estimation methods leverage commonly available sensors typically found on autonomous platforms. Several studies \cite{GNSS_SLIP, GNSS_SLIP2, GNSS_SLIP3, GNSS_SLIP4, GNSS_DOUBLE_ANTENNA_SLIP} combine inertial measurement unit (IMU) data with GNSS sensors, either single or dual antenna, to estimate slip angle. However, these approaches present two critical limitations. First, in GNSS-denied environments, velocity measurements become unavailable, and reliance solely on IMU acceleration could lead to rapid estimator divergence due to noise and bias integration. Second, the relatively low rate of GNSS measurements may be insufficient to capture fast changing dynamics typical of high-speed scenarios, such as autonomous racing.

Another research direction focuses on camera-based methods that emulate optical sensors for vehicle dynamics estimation. These techniques \cite{CAMERA_SLIP_DOWNFACE, CAMERA_SLIP_DOWNFACE2, CAMERA_SLIP_DOWNFACE3, CAMERA_SLIP_CORRELATION, CAMERA_SLIP_RC_SCALED} use downward-facing cameras to track salient ground features between frames. By leveraging known camera intrinsics and extrinsics, and assuming that all tracked features lie on the ground plane, pixel displacements can be converted into velocity vectors. However, these methods require precise camera placement and textured road surfaces. Additionally, at high vehicle speeds, cameras must operate at very high frame rates to ensure feature overlap between consecutive frames. This, in turn, necessitates a high-performance processing platform capable of handling high data throughput and performing real-time image processing.

Additional sensors such as LiDAR can be leveraged to estimate vehicle lateral dynamics. In our previous work, LOP-UKF \cite{LOP_UKF}, we utilized LiDAR-based localization to obtain lateral velocity measurements, which were then fused with a single-track vehicle model using an Unscented Kalman Filter (UKF). The lateral tire forces were modeled using the Pacejka Magic Formula, with parameters estimates obtained through offline calibration based on experimental data. While effective, this method relies on the prior collection of LiDAR data to reconstruct a high-fidelity track map for accurate localization, as well as precomputed tire model parameters.

In recent years, millimeter-wave frequency-modulated continuous wave (mmWave FMCW) RADAR has attracted significant attention for localization and navigation applications. RADAR sensors provide not only range and bearing information but also direct measurements of Doppler velocity. In environments with predominantly static objects, the ego-velocity of the sensor platform can be estimated from these Doppler measurements. Frameworks such as CREVE \cite{CREVE} and RAVE \cite{RAVE} exploit this to robustly estimate ego-vehicle motion by filtering out outliers caused by noise or dynamic objects. This is achieved through a combination of outlier rejection techniques and robust estimation algorithms. The resulting velocity estimates can be seamlessly integrated into graph-based localization systems \cite{GARLIO, ROAMER, LIRO}, and as shown in \cite{RVF}, the addition of these measurements significantly enhances both longitudinal and lateral velocity estimation accuracy. 
Based on this, an Adaptive Kalman Filter is used in \cite{RADAR_SIDE_SLIP_FILTER} to estimate the vehicle velocity and the slip gradient by integrating RADAR Doppler measurements into a motion observer.

While RAGE adopts a similar approach, it is specifically designed to address a different set of challenges that arise in the context of autonomous racing:
\begin{itemize}
    \item RADAR Doppler measurements can be subject to non-negligible delays due to internal sensor processing. If these delays are not properly compensated, they may cause inconsistencies between the velocity corrections and the measured linear accelerations, in particular during rapid acceleration or deceleration events. To address this, we adopt MHE approach, which is well-suited for handling delayed or asynchronous measurements, provided the measurement timestamps fall within the current estimation horizon.
    \item Due to the nature of RADAR sensor signal processing \cite{RADAR_ALIAS_DESC}, only velocities within a specific range can be directly measured. Velocities outside this range result in aliasing effects, where the measured velocity is misrepresented. In the context of autonomous racing, where vehicle speeds often exceed this measurable range, aliasing becomes inevitable. To address this, we integrated into our estimator a simple de-aliasing approach adapted from methods used in operational meteorology in order to recover the true Doppler velocity measurements.
    \item RAGE employs a non-linear single-track model, as many of the linear approximations used in \cite{RADAR_SIDE_SLIP_FILTER} become invalid at the high velocities and accelerations. Furthermore, RAGE performs online fitting of a non-linear tire model, enabling a more accurate representation of the lateral forces acting on both axles.
\end{itemize}


\section{System Overview} \label{sec:overview}
The RAGE estimator is formulated and solved as a Moving Horizon Estimation (MHE) problem. The objective is to minimize the sum of the residuals from prior information, system dynamics and available measurements to obtain the maximum posterior estimate of the states of the ego vehicle and tire parameters.
The variables to be estimated are:
\begin{align}
    \boldsymbol{\Omega} = [ \boldsymbol{X}, \boldsymbol{P} ]
\end{align}
where $\boldsymbol{X}$ represents the sequence of vehicle states over a sliding time window of length $N+1$:
\begin{align}
    &\boldsymbol{X} = [ \mathbf{x}_{0}, \dots, \mathbf{x}_{N} ]\\ \quad
    &\mathbf{x}_{k} = [ v_{x_k}, v_{y_k}, r_k, b_{x_k}, b_{y_k}, b_{r_k} ], \quad k \in [0, N]
\end{align}
Here, $v_x$ and $v_y$ are the longitudinal and lateral velocities of the vehicle, $r$ is the yaw rate of the vehicle and $b_x$, $b_y$, and $b_r$ are the IMU biases in the longitudinal and lateral accelerations and the measurements of the yaw rate, respectively.

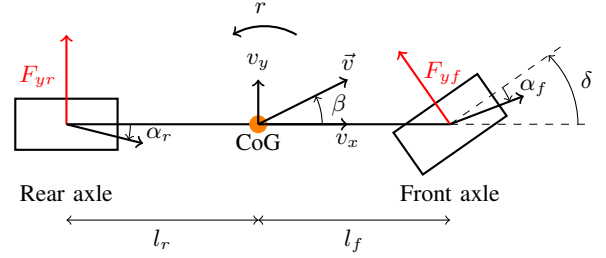
\begin{figure}[b]
  \centering
  \begin{tikzpicture}[scale=1.7, every node/.style={font=\small}]

    \draw[thick] (2, 2) -- (5, 2) node[midway, above] {};

    \draw[thick] (2-0.4, 2-0.2) rectangle (2+0.4, 2+0.2);
    \node[below] at (2, 1.6) {Rear axle};

    \draw[->, thick, red] (2, 2) -- (2, 2.7) node[pos=0.3, above left] {$F_{yr}$};  

    \draw[->, thick, black] (2, 2) -- (2.6, 1.85);
    \draw[->] (2, 2) ++(0:0.5) arc[start angle=0, end angle=-14, radius=0.5]; 
    \node[right] at (2.55, 1.92) {$\alpha_r$}; 

    \begin{scope}[shift={(5, 2)}] 
        \begin{scope}[rotate=35]
            \draw[thick] (-0.4, -0.2) rectangle (0.4, 0.2);
            \draw[->, thick, red] (0, 0) -- (0, 0.7) node[pos=0.7, above right, right=0] {$F_{yf}$};  

            \draw[ultra thin, dashed] (0, 0) -- (1.1, 0);

            \draw[->, thick, black] (0, 0) -- (0.6, -0.15);
            \draw[->] (0.0, 0) ++(0:0.5) arc[start angle=0, end angle=-14, radius=0.5]; 
            \node[right] at (0.55, -0.04) {$\alpha_f$}; 
        \end{scope}

        \draw[->] (0.5, 0) ++(0:0.5) arc[start angle=0, end angle=46, radius=0.75]; 
        \node[right] at (0.95, 0.35) {$\delta$}; 
        \draw[ultra thin, dashed] (0, 0) -- (1.1, 0);
    \end{scope}
    \node[below] at (5, 1.6) {Front axle};


    \fill[orange] (3.5, 2) circle (2pt);
    \node[below] at (3.5, 2.0) {CoG};

    \draw[<->] (2, 1.25) -- (3.5, 1.25) node[midway, below] {$l_r$}; 
    \draw[<->] (3.5, 1.25) -- (5, 1.25) node[midway, below] {$l_f$}; 

    \begin{scope}[shift={(3.5, 2)}] 
    \draw[->, thick] (0, 0) -- (0.7, 0.35) node[anchor=south] {$\vec{v}$};
    \draw[->, thick] (0, 0) -- (0.7, 0) node[anchor=north] {$v_x$};
    \draw[->, thick] (0, 0) -- (0, 0.35) node[anchor=south] {$v_y$};
    \draw[->] (0, 0) ++(0:0.5) arc[start angle=0, end angle=26, radius=0.5]; 
        \node[right] at (0.5, 0.15) {$\beta$}; 
    \end{scope}

    \draw[<-, thick] (3.0, 2.4) ++(0.3, 0.3) arc[start angle=120, end angle=60, radius=0.5];
    \node[above right] at (3.4, 2.8) {$r$}; 
  \end{tikzpicture}
  \caption{Schematic representation of the dynamic single-track model.}\label{fig:bicycle}
\end{figure}

\nic{In the proposed framework, a IMU-driven 3 DOF (Degree of Freedom) vehicle model is used for the estimation of the vehicle motion (Sec. \ref{sec:model}), while a nonlinear dynamic single-track model (Figure \ref{fig:bicycle}) is adopted for tire force estimation (Sec. \ref{sec:lateral}).}
The relationship between the slip angles, the vertical load on each axle, and the corresponding lateral force is characterized using the Pacejka Magic Formula \cite{PACEJKA}. The set of macro-parameters that define the Pacejka tire model for both the front and rear axles is denoted by $\boldsymbol{P}$:
\begin{align}
    &\boldsymbol{P} = [ \mathbf{p}_{f}, \mathbf{p}_{r} ] \\
    &\mathbf{p}_{j} = [ B_{j}, C_{j}, D_{j}, E_{j}, S_{hj}, S_{vj} ], \quad j \in [f, r]
\end{align}
The subscript $j$ indicates whether the parameters apply to the front ($f$) or rear ($r$) axle.

The system input vector at time step $k$ is:
\begin{align}
    \mathbf{u}_k = [\hat{a}_{x_k}, \hat{a}_{y_k}, \hat{r}_k, \hat{\delta}_k] , \quad k \in [0, N]
\end{align}
These correspond to the longitudinal acceleration $\hat{a}_x$, the lateral acceleration $\hat{a}_y$ and the yaw rate $\hat{r}$ measured by the IMU and the angle of the steering wheel $\hat{\delta}$ obtained from the vehicle’s feedback.

Concisely, the RAGE problem can be defined as follows
\begin{align}
\underset{\Omega}{\mathrm{min}} \quad
&\|\mathbf{x}_0 - \tilde{\mathbf{x}}_0\|^2_{\Sigma_{\mathbf{x}_0}} + 
 \|\mathbf{P} - \tilde{\mathbf{P}}\|^2_{\Sigma_{\mathbf{P}}} \\
&+ \sum_{k=0}^{N} \left( 
    \|\mathbf{w}_k\|^2_{\Sigma_{\mathbf{w}_k}} + 
    \sum_{l=0}^{M} \|\mathbf{v}^l_k\|^2_{\Sigma_{\mathbf{v}^l_k}} 
\right)
\end{align}
subject to
\begin{align}
\mathbf{x}_{k} &= f(\mathbf{x}_{k-1}, \mathbf{u}_{k-1}) + \mathbf{w}_k \qquad k = 1 \dots N \\
\mathbf{\hat{y}}^l_k &= h_l(\mathbf{x}_k, \mathbf{u}_k, \mathbf{P}) + \mathbf{v}^l_k \quad k = 0 \dots N \quad l = 0 \dots M \\
&\mathbf{P}_{\min} \leq \mathbf{P} \leq \mathbf{P}_{\max}
\end{align}

The MHE optimization problem is solved iteratively. When the estimation window shifts forward, the initial state becomes the second state from the previous window, a new state is added to the end of the window, and the most recent measurements are incorporated while discarding the oldest ones.
The value of \( \tilde{\mathbf{x}}_0 \) serves as a prior for the initial state of the current window. It is derived from the corresponding state in the previous estimate to maintain consistency across iterations. The associated covariance matrix \( \Sigma_{\mathbf{x}_0} \) regulates how much the new estimate is allowed to deviate from this prior, ensuring smoothness and continuity in the evolution of the states.
Similarly, \( \tilde{\mathbf{P}} \) represents the prior for the parameters \( \mathbf{P} \), and is updated at each step using the parameters estimate from the previous iteration. The covariance matrix \( \Sigma_{\mathbf{P}} \) controls how quickly the parameters can adapt to new measurements, effectively determining how much of the past information is retained versus forgotten.
The function \( f(\cdot) \) describes the system's non-linear state transition model, predicting the next state \( \mathbf{x}_{k+1} \) based on the current state \( \mathbf{x}_k \) and control inputs \( \mathbf{u}_k \).
The set of functions \( \{ h_l(\cdot) \}_{l=0}^{M} \) represents the measurement models, which map the current state, inputs, and parameters to the measured sensor outputs \( \mathbf{\hat{y}}_k^l \).
Both the process and measurement models are assumed to be affected by additive zero-mean Gaussian noises, respectively, process noise \( \mathbf{w}_k \sim \mathcal{N}(0, \Sigma_{\mathbf{w}_k}) \) and measurement noise \( \mathbf{v}_k^l \sim \mathcal{N}(0, \Sigma_{\mathbf{v}_k^l}) \).
Lastly, the parameters \( \mathbf{P} \) are constrained within predefined bounds to ensure physical plausibility.

In RAGE, a new state is added to the sliding window every 10 ms to maintain high-resolution tracking of vehicle dynamics. However, the full MHE optimization problem is solved only when valid RADAR measurements are received and incorporated into the estimation process. This approach ensures that the state estimates are corrected using the most recent velocity observations, while also allowing the computationally intensive optimization to run less frequently than the state update rate. RADAR measurements are typically available every 20 ms, and each optimization step is triggered accordingly. The sliding window is shifted only after an optimization step is completed, until the time span between the oldest and newest states in the window becomes shorter than the maximum allowed horizon length, denoted as \( \Delta T_w = 150 \) ms. During each window shift, the prior for the initial state \( \tilde{\mathbf{x}}_0 \) and the parameter vector \( \tilde{\mathbf{P}} \) are updated to reflect the latest estimates from the previous optimization step.

\section{Motion Model}\label{sec:model}
The RAGE estimator utilizes a \nic{IMU-driven, rigid-body} model in curvilinear motion. In this model, longitudinal and lateral velocities and the yaw rate serve as state variables, while longitudinal and lateral accelerations are treated as system inputs. To enhance estimation accuracy, IMU measurements biases are also estimated. 
The yaw rate $r_k$, measured directly by the IMU, is corrected by subtracting the estimated bias to produce an unbiased angular velocity. All IMU biases are modeled as random walk processes, allowing them to evolve gradually over time to account for sensor drift. Based on this formulation, the system state transition model is defined as follows:
\begin{align}
    &\mathbf{x}_{k} = f(\mathbf{x}_{k-1}, \mathbf{u}_{k-1}) \\
    &\left\{ 
    \begin{array}{l}
    v_{x_{k}} = v_{x_{k-1}} + \left(\hat{a}_{x_{k-1}} - b_{x_{k-1}} + r_{k-1} v_{y_{{k-1}}}\right) \Delta t  \\
    v_{y_{k}} = v_{y_{k-1}} + \left(\hat{a}_{y_{k-1}} - b_{y_{k-1}} - r_{k-1} v_{x_{{k-1}}}\right) \Delta t \\
    r_{k} = \hat{r}_{k-1} - b_{r_{k-1}} \\
    b_{x_{k}} =  b_{x_{k-1}} \\
    b_{y_{k}} =  b_{y_{k-1}} \\
    b_{r_{k}} =  b_{r_{k-1}} \\
    \end{array}
    \right.
\end{align}
The forward Euler method was selected as the integration scheme because of its simplicity and computational efficiency compared to higher-order methods. Despite its basic formulation, it has demonstrated stable behavior in all experimental scenarios. The chosen integration step size is \( \Delta t = 10 \) ms, which offers a good balance between numerical stability and real-time performance.

\section{IMU Biases Initialization}\label{sec:bias}
The initialization of IMU biases is performed when the vehicle is stationary. The vehicle is considered to be at rest if the following two conditions hold continuously for at least \( T_{\text{stop}} \) seconds:
\begin{itemize}
\item the total absolute velocity is below \( V_{\text{min}} = 0.5 \ \text{m/s} \)
\item the gravity-compensated acceleration magnitude is less than \( A_{\text{min}} = 0.2 \ \text{m/s}^2 \)
\end{itemize}
Once both conditions are satisfied, the IMU measurements from the last \( T_{\text{stop}} \) seconds are used to calculate the vehicle's 3D attitude using a Madgwick orientation filter \cite{MADGWICK}. This filter provides the rotation matrix \( \mathbf{R}_W^B \), which transforms the gravity vector from the world frame $W$ to the body frame $B$:
\begin{align}
\mathbf{g}^B = \mathbf{R}_W^B \mathbf{g}^W
\end{align}
Then, we compensate the measured accelerations for gravity:
\begin{align}
\tilde{a}_x &= \hat{a}_x - g_x^B \\
\tilde{a}_y &= \hat{a}_y - g_y^B
\end{align}
Since the vehicle is known to be motionless during this phase, the velocity, linear acceleration, and angular velocity are assumed to be zero. This enables the application of the Zero-Velocity Update (ZUPT) method \cite{ZUPT}, by introducing the following measurement model into the MHE formulation:
\begin{align}
&\mathbf{\hat{y}}^{ZV}_k = h_{ZV}(\mathbf{x}_k, \mathbf{u}_k) \\
&\left\{
\begin{array}{l}
0 = v_{x_k}  \\
0 = v_{y_k}  \\
0 = r_k \\
\tilde{a}_{x_k} = b_{x_k} \\
\tilde{a}_{y_k} = b_{y_k} \\
\hat{r}_k = b_{r_k}
\end{array}
\right.
\end{align}
These conditions allow RAGE to directly observe and correct the IMU biases during stationary periods.

\section{Doppler Velocity Measurements} \label{sec:radial}
Each RADAR point cloud consists of a set of 3D points expressed in polar coordinates \((\hat{r}, \hat{\theta}, \hat{\phi})\) representing range, azimuth and elevation, along with their corresponding Doppler velocities \(\hat{v}_d\). Each point also includes a Signal-to-Noise Ratio (SNR) value. Filtering out RADAR measurements with low SNR has been observed to produce better velocity estimates.

Due to internal processing time, RADAR measurements are received with non-negligible latency. These delays can be measured using time synchronization protocols such as gPTP \cite{PTP}. To accurately associate each measurement to the correct state, rather than associating it with the most recent state, we insert a new state back in time corresponding to the timestamp when the RADAR data was actually captured.

Assuming a point is stationary relative to the vehicle, its measured Doppler velocity can be modeled as:
\begin{align} \label{eq:radial_vel}
\nic{v_d = -\mathbf{b}(\theta, \phi)^\top \mathbf{v}_R^R}
\end{align}
where \(\mathbf{b}(\theta, \phi)\) is the unit bearing vector from the RADAR to the target point, and \(\mathbf{v}_R^R\) is the translational velocity of the RADAR sensor expressed in its own frame.
The RADAR velocity \(\mathbf{v}_R^R\) is related to the velocity of the vehicle \(\mathbf{v}_B^B\) as:
\begin{align} \label{eq:radar_vel}
\nic{\mathbf{v}_R^R = \mathbf{R}_R^{B^\top} \left( \mathbf{v}_B^B + \boldsymbol{\omega}_B^B \times \mathbf{t}_R^B \right)}
\end{align}
Here, \(\boldsymbol{\omega}_B^B\) is the vehicle's angular velocity vector,
\(\mathbf{t}_R^B\) is the position of the RADAR in the body frame, and \(\mathbf{R}_R^B\) is the rotation matrix from the RADAR frame to the body frame.

The measured Doppler velocity \( \hat{v}_d \) from a RADAR is actually an apparent value, not necessarily the true Doppler velocity \( \hat{v}_r \). Due to the nature of RADAR signal processing \cite{RADAR_ALIAS_DESC}, the sensor can only unambiguously measure Doppler velocities within a limited range defined by the Nyquist velocity \( V_N \). Any true Doppler velocity outside the interval \([ -V_N, V_N ]\) will be aliased back into this range.
The relationship between the true Doppler velocity \( \hat{v}_r \) and the measured velocity \( \hat{v}_d \) is given by the following:
\begin{align} \label{eq:dealias}
    \hat{v}_r = \hat{v}_d + 2 n V_N
\end{align}
where \( n \in \mathbb{Z} \) is an unknown integer that represents how many times the true velocity has been "wrapped around" the Nyquist limit.
This phenomenon is known as velocity aliasing, and the process of recovering \( \hat{v}_r \) from \( \hat{v}_d \) is called de-aliasing.
A practical and efficient way to perform de-aliasing \cite{DEALIAS} is to use an expected Doppler velocity \( v_e \). Using this prediction, the integer \( n \) can be estimated as:
\begin{align}
    n = \mathrm{nint}\left(\frac{v_e - \hat{v}_d}{2V_N}\right)
\end{align}
where \(\mathrm{nint}(\cdot)\) denotes the nearest integer function. Once \( n \) is known, the corrected Doppler velocity can be computed by (\ref{eq:dealias}).
The expected Doppler velocity \( v_e \) is calculated from the state of the vehicle at the time of the RADAR measurement, using Equations (\ref{eq:radial_vel}) and (\ref{eq:radar_vel}). This allows for robust and accurate dealiasing based on the current velocity estimate.

Finally, the resulting RADAR measurement model for an individual point \(i\) at time step \(k\) is:

\begin{align}
\mathbf{\hat{y}}^R_k &= h_R(\mathbf{x}_k) \\
\hat{v}^i_{r_k} &= -\mathbf{b}(\hat{\theta}^i_k, \hat{\phi}^i_k)^\top \left( \mathbf{R}_R^{B^\top} \left( 
\begin{bmatrix}
v_{x_k} \\
v_{y_k} \\
0
\end{bmatrix} 
+ 
\begin{bmatrix}
0 \\
0 \\
r_k
\end{bmatrix} 
\times \mathbf{t}_R^B 
\right)\right)
\end{align}

Assuming low elevation angles and negligible vertical motion, we simplify the model by ignoring the z-component velocity and the angular velocities around the x and y axes.

Before incorporating a RADAR point into the MHE problem, we compute the initial measurement model error of the Doppler velocity. If this error exceeds a threshold \(\Delta V_r^{\text{max}}\), the measurement is discarded as a likely outlier (e.g., due to ghosting effects or moving objects).

To further reduce the sensitivity of the estimation process to such outliers, a Cauchy loss function is applied to the residuals from RADAR measurements during optimization. This robust loss mitigates the impact of erroneous points while preserving the influence of valid observations.

\section{Lateral Dynamics} \label{sec:lateral}

\nic{While the IMU-driven 3 DOF model is sufficient for motion estimation, it does not explicitly capture the underlying force interactions. For this reason, a nonlinear dynamic single-track model is adopted for tire model estimation, allowing to represent the force balance between the front and rear axles.
Tire forces are modeled using a nonlinear Pacejka Magic Formula \cite{PACEJKA}}. To compute these lateral forces, we first determine the slip angles for the front and rear wheels based on the current state of the vehicle \(\mathbf{x}_k\). The front slip angle \(\alpha_{f_k}\) and the rear slip angle \(\alpha_{r_k}\) are computed as:
\begin{align}  
\alpha_{f_k} &= \arctan\left(\frac{v_{y_k} + r_k \, l_f}{v_{x_k}}\right) - \hat{\delta}_k
\\
\alpha_{r_k} &= \arctan\left(\frac{v_{y_k} - r_k \, l_r }{v_{x_k}}\right)
\end{align}
where the vehicle velocity components $v_{x_k}$ and $v_{y_k}$  are transformed into the front and rear axle frames, taking into account the transformation due to the distances from the vehicle's center of gravity, $l_f$ and $l_r$, as well as the rotation due to the steering angle $\hat{\delta}_k$.

The vertical loads on each wheel, \(F_{zf}\) and \(F_{zr}\), are composed of several contributing factors: a static load determined by the vehicle's weight distribution, a dynamic load transfer caused by longitudinal acceleration, and aerodynamic downforce \nic{\cite{guiggiani2023science}}. Since the employed single-track model does not differentiate between the effects of the left and right tires during cornering, the effect of lateral load transfer is intrinsically incorporated into the resulting axle characteristics. The vertical loads are then modeled as follows:
\begin{align}
F_{zf}(\mathbf{x}_k, \mathbf{u}_k) = \frac{m}{l_f + l_r} \left( g l_r - \hat{a}_x h_g \right) + \frac{1}{2} C_{zf} \rho v_x^2 A
\\
F_{zr}(\mathbf{x}_k, \mathbf{u}_k)= \frac{m}{l_f + l_r} \left( g l_f + \hat{a}_x h_g \right) + \frac{1}{2} C_{zr} \rho v_x^2 A
\end{align}
where \(m\) is the total vehicle mass, \(h_g\) is the height of the CoG, \(g\) is the gravitational acceleration, $A$ is the projected area of the vehicle, and $C_{zf}$ and $C_{zr}$ denote the front and rear downforce coefficients, respectively.

Using the slip angles and the vertical loads, the front and rear lateral tire forces are computed as:
\begin{align}\label{eq:Fy}    
F_{yf_k}(\mathbf{x}_k, \mathbf{u}_k, \mathbf{p}_{f}) &= F_{zf}(\mathbf{x}_k, \mathbf{u}_k) Y(\alpha_{f_k}, \mathbf{p}_{f})\\
F_{yr_k}(\mathbf{x}_k, \mathbf{u}_k, \mathbf{p}_{r}) &= F_{zr}(\mathbf{x}_k, \mathbf{u}_k) Y(\alpha_{r_k}, \mathbf{p}_{r}) 
\end{align}
where \(Y(\cdot)\) is the Pacejka Magic Formula:
\begin{align}\label{eq:pacejka}
y(x, \mathbf{p}) &= D \sin( C \atan (B x - E (B x - \atan(Bx)))) \\
Y(x, \mathbf{p}) &= y(x + S_h, \mathbf{p}) + S_v
\end{align}

\nic{Note that, according to Equation \eqref{eq:pacejka}, the peak factor $D$ determines the maximum lateral axle force \cite{PACEJKA,guiggiani2023science}.
Therefore, estimating this parameter provides direct information on the maximum lateral forces that each axle can generate, and thus on the available grip level.
}

For rear-wheel-drive racing car application, and under the assumption of small steering angles, the effect of longitudinal forces on lateral dynamics is neglected to simplify the analysis, although it may not be entirely negligible. Therefore, the measured lateral forces on the front and rear axles can be obtained via inertial measurements as follows:
\begin{align}
\hat{F}_{yf_k} = \frac{l_r}{l_f+l_r} \frac{m \hat{a}_{y_k}}{\cos \hat{\delta}} \label{eq:fyf}\\
\hat{F}_{yr_k} =  \frac{l_f}{l_f+l_r} m \hat{a}_{y_k} \label{eq:fyr}
\end{align}

Combining the model-based and measurement-based formulations, the lateral force measurement model is expressed compactly as:
\begin{align}
&\mathbf{\hat{y}}_k^{F_y} = h_{F_y}(\mathbf{x}_k, \mathbf{u}_k, \mathbf{P})
\\
&\left\{ 
\begin{array}{l}
\hat{F}_{yf_k} = F_{zf}(\mathbf{x}_k, \mathbf{u}_k) Y(\alpha_{f_k}, \mathbf{p}_{f})\\
\hat{F}_{yr_k} = F_{zr}(\mathbf{x}_k, \mathbf{u}_k) Y(\alpha_{r_k}, \mathbf{p}_{r}) 
\end{array}
\right.
\end{align}
To ensure the reliability of the measurement model, it is included in the MHE formulation only when the total velocity of the vehicle at state \(\mathbf{x}_k\) exceeds a predefined threshold \(V_{F_y}^{\text{min}}\). This constraint is made because at low longitudinal speeds the lateral tire forces become negligible, and also the slip angle calculations approach a singularity due to the division by small values of \(v_x\), leading to numerical instability.
\section{Validation} \label{sec:simulation}
\begin{figure}[b]
\centering
\includegraphics[width=0.85\linewidth]{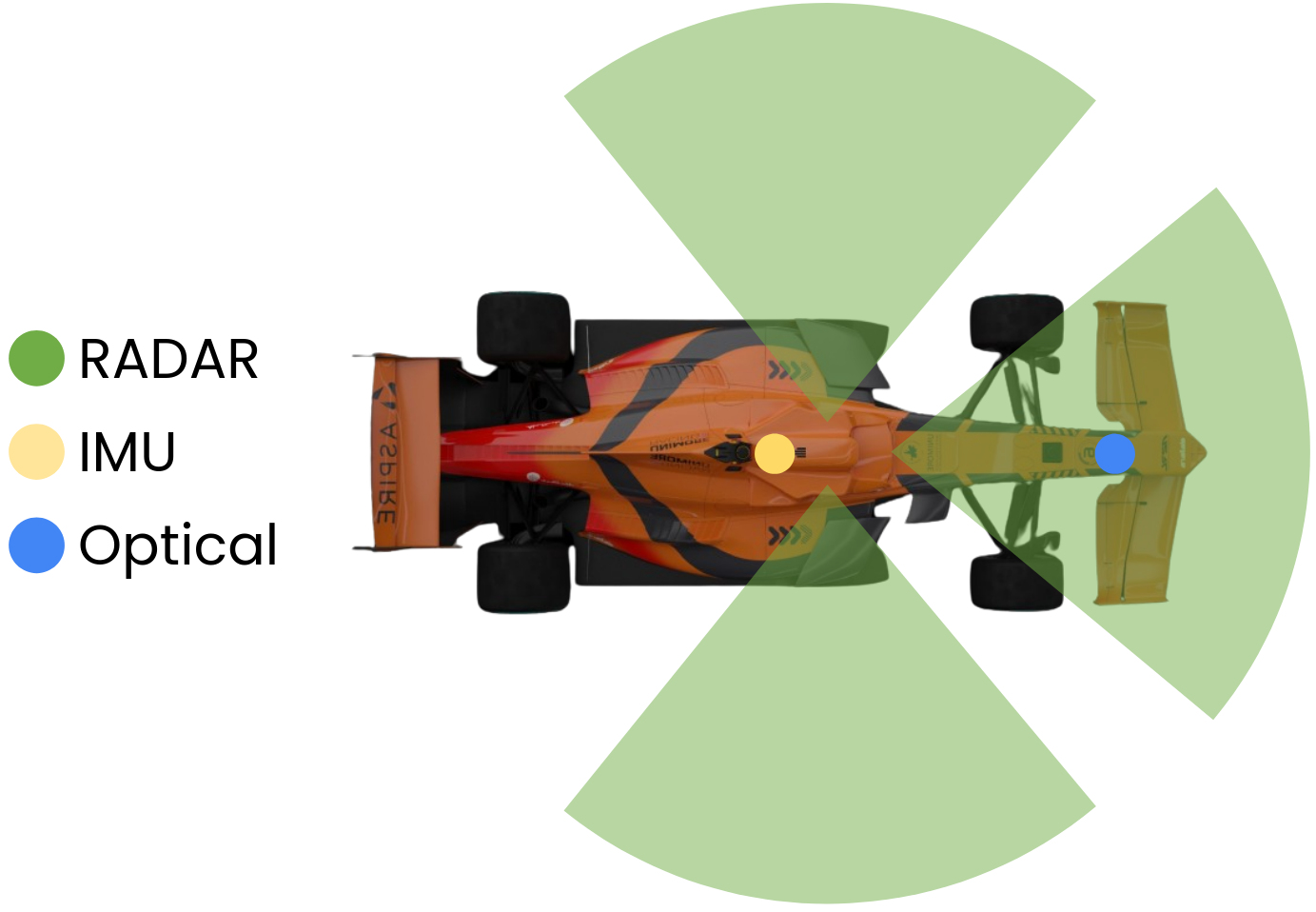}
\caption{Sensor setups for the Dallara EAV-24. While the car is equipped with
additional sensors, only the ones used in this study are shown.}
\label{fig:sensor}
\end{figure}

To validate the RAGE estimator, we first run different tests in simulations using our multi-body vehicle simulator \cite{ER_AUTOPILOT}. This simulator provides accurate ground-truth data for the vehicle's body-frame velocity $\mathbf{v}_B^B$, accelerations, and the lateral and vertical forces acting on each axle. Sensor measurements are simulated as if originating from the sensor setup depicted in Figure \ref{fig:sensor}. The EAV-24 autonomous race car is equipped with a comprehensive suite of sensors. These include a steering angle sensor, an inertial measurement unit (IMU), the VectorNav VN-310, supplies acceleration and angular rate data at a sampling frequency of 200 Hz and three ZF ProWave RADAR sensors, one mounted at the front and two on the lateral sides. Each RADAR sensor delivers a 4D point cloud at a native frequency of 16.6 Hz and with a maximum Doppler velocity of 26.5 m/s. To ensure precise temporal alignment, all RADARs are synchronized with the onboard computer system using gPTP protocol \cite{PTP}, enabling highly accurate timestamping of each individual measurement. The typical observed acquisition latency for these RADARs ranges between 80 ms and 100 ms. Also, each RADAR is triggered in a predefined time slot, allowing for staggered acquisitions across the three sensors. Specifically, a 20 ms time offset is introduced between each RADAR trigger. This configuration enables the system to effectively emulate a single RADAR operating at nearly 50 Hz, thereby achieving a higher temporal resolution. Additionally, the vehicle is equipped with a Kistler SF-Motion optical sensor, which employs non-contact technology to directly measure both longitudinal and lateral velocities of the vehicle with high precision at 100 Hz.

In the simulator, IMU data is synthetically generated by adding white Gaussian noise to the simulator’s reference linear acceleration and angular velocity signals. The noise characteristics are tuned to match those observed from a real IMU mounted on the race car. Similarly, synthetic RADAR data is generated. Since we are only interested in the radial information from the RADAR, rather than a detailed environmental point cloud, we only generate azimuth, elevation and Doppler velocity $(\hat{\theta}_i, \hat{\phi}_i, \hat{v}_{d_i})$.
For each RADAR point cloud, a random number of points $N \sim \mathcal{N}(\mu_N,\sigma^{2}_N)$ is generated. The process to generate each point is defined as follows:
\begin{align}
&\theta_i \sim \text{Cauchy}(\mu_{\theta}, \gamma_\theta) \\
&\phi_i \sim \text{Cauchy}(\mu_{\phi}, \gamma_\phi) \\
&v_{r_i} = -\mathbf{b}(\theta_i, \phi_i)^\top \left( \mathbf{R}_R^{B^\top} \left( \mathbf{v}_B^B + \boldsymbol{\omega}_B^B \times \mathbf{t}_R^B \right) \right)\\
&\hat{\theta}_i = \theta_i + n_\theta \hspace{2.6cm} n_\theta \sim \mathcal{N}(0, \sigma^2_\theta) \\
&\hat{\phi}_i = \phi_i + n_\phi \qquad\qquad\qquad\quad n_\phi \sim \mathcal{N}(0, \sigma^2_\phi) \\
&\hat{v}_{d_i} = \text{wrap}(v_{r_i} + n_{v_d}, V_N)  \quad\, n_{v_d} \sim \text{Cauchy}(\mu_{v_d}, \phi_{v_d})
\end{align}
The ground-truth azimuth $\theta_i$ and elevation $\phi_i$ are sampled independently from two Cauchy distributions. The true Doppler velocity $v_{r_i}$ is computed by projecting the vehicle's velocity vector $\mathbf{v}_B^B$ onto the obtained bearing vector. The final azimuth $\hat{\theta}_i$ and elevation $\hat{\phi}_i$ are obtained by adding white Gaussian noise to the corresponding ground-truth angles to account for measurement uncertainty. To simulate Doppler velocity measurements, the function $\text{wrap}(\cdot)$ is used to model the aliasing effect, wrapping the true Doppler velocity $v_{r_i}$ plus some noise $n_{v_d}$ into the observable Doppler velocity $\hat{v}_{d_i}$, constrained by the Nyquist velocity limit $V_N$.
Finally, the generated point cloud is emitted with a time delay $t_d \sim \mathcal{N}(\mu_{t_d}, \sigma^2_{t_d})$, representing internal processing latency within the RADAR. For simplicity, all random variables and noise sources are assumed to be mutually independent, and the parameters of the distributions are estimated from real-world data.


\subsection{Double-Lane Change Maneuver}
\begin{figure}[b]
    \vspace{-0.45cm}
    \centering
    \includegraphics[width=\linewidth]{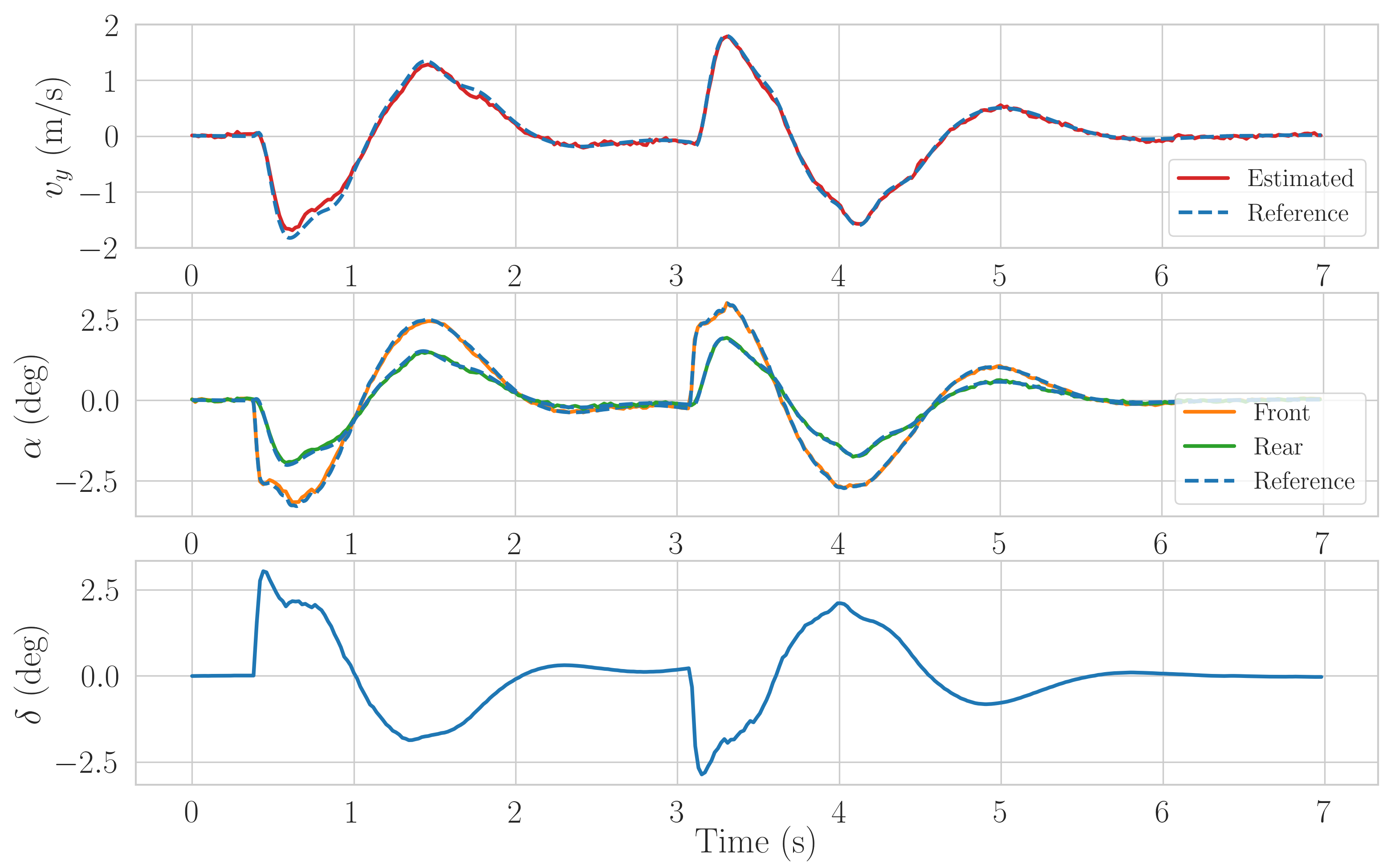}
    \caption{\nic{Simulated double lane change maneuver. The simulator outputs slip angles and lateral forces for all four tires. For the single-track model, slip angles are averaged and forces summed per axle.}}
    \label{fig:sim_dlc}
\end{figure}

The first simulated test evaluates a double-lane change maneuver performed according to the ISO 3888-1 standard \cite{ISO3888-1}. This test is conducted on dry asphalt with a friction coefficient of $\mu = 0.8$, at a longitudinal speed of 65 m/s, simulating a high-speed obstacle avoidance scenario. Figure \ref{fig:sim_dlc} illustrates the estimation results for this case. The data demonstrate that RAGE maintains high accuracy in tracking rapid and substantial changes in lateral velocity and slip angles, even under aggressive maneuvers characterized by high lateral acceleration. The maximum absolute error in estimated lateral velocity is 0.19 m/s, while the maximum absolute error in estimated slip angle is 0.15°.

\subsection{Spin}
\begin{figure}[b]
    \centering
    \includegraphics[width=\linewidth]{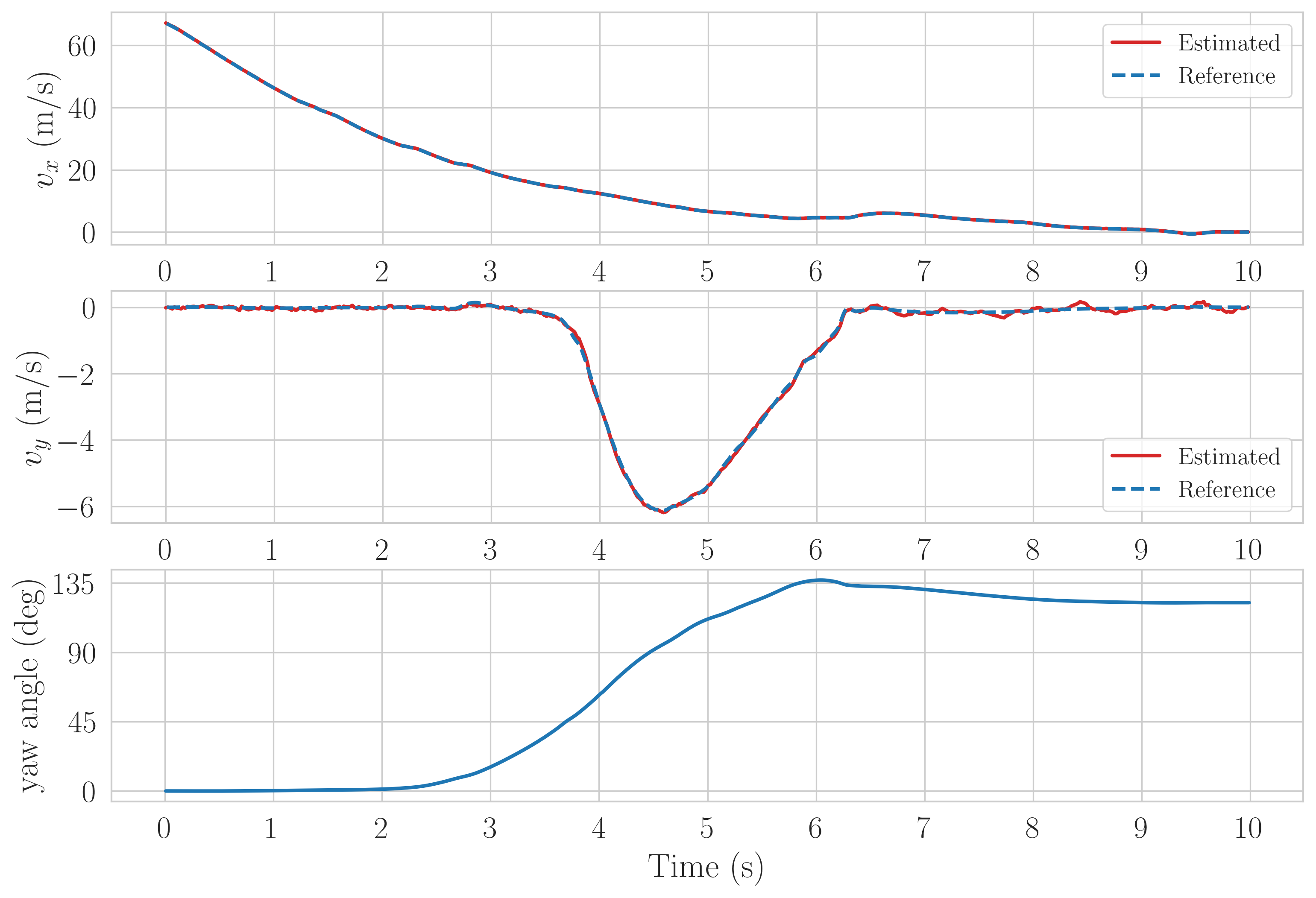}
    \caption{Simulated spin on wet asphalt ($\mu = 0.6$), the car brakes from 65 m/s before a left turn, but reduced grip causes rear instability and a 135° spin.}
    \label{fig:sim_spin}
\end{figure}

In the second simulated test, the car accelerates to 65 m/s on a straight section of wet asphalt ($\mu = 0.6$). Upon reaching the end of the straight, the vehicle initiates braking to approach a left turn. As a result of the reduced tire-road friction, the vehicle becomes unstable during the deceleration phase. This instability causes the rear of the car to lose traction, leading to a spin of approximately 135°, as depicted in Figure \ref{fig:sim_spin}. Despite these complexities, RAGE successfully maintains accurate state tracking throughout the maneuver. Notably, it achieves this using a simple motion model, demonstrating its robustness and reliability even in highly non-linear dynamic situations. 

\begin{figure}[tb]
    \centering
    \includegraphics[width=\linewidth]{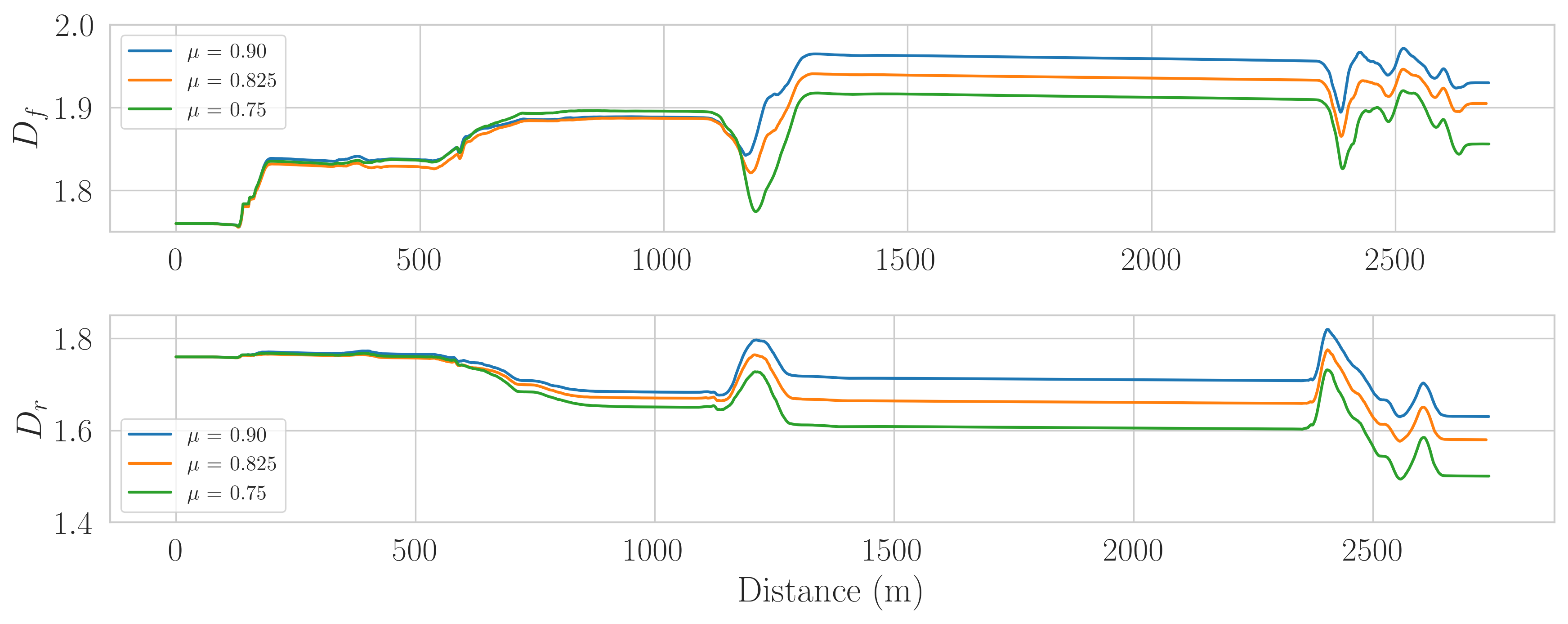}
    \caption{\nic{Estimated Pacejka peak coefficient $D$ under three different road friction levels.}}
    \label{fig:sim_d_peak}
\end{figure}

\nic{
\subsection{Tire-Road Friction}
In the third simulation experiment, we assess the estimated peak factor $D$ under different tire–road friction levels ($\mu$). Across all tests, the vehicle follows the same reference trajectory, while only $\mu$ is varied to emulate different surface conditions. As illustrated in Fig.~\ref{fig:sim_d_peak}, the parameter $D$ decreases consistently with $\mu$, reflecting the reduction in the maximum achievable axle force according to the adopted Pacejka model.
This behavior supports the interpretation of $D$ parameter as a model-based indicator of the available tire–road grip level and shows that the proposed estimator can effectively capture variations in the underlying friction conditions.
}

\subsection{Parameters Fitting}
\begin{figure}[tb]
    \centering
    \includegraphics[width=\linewidth]{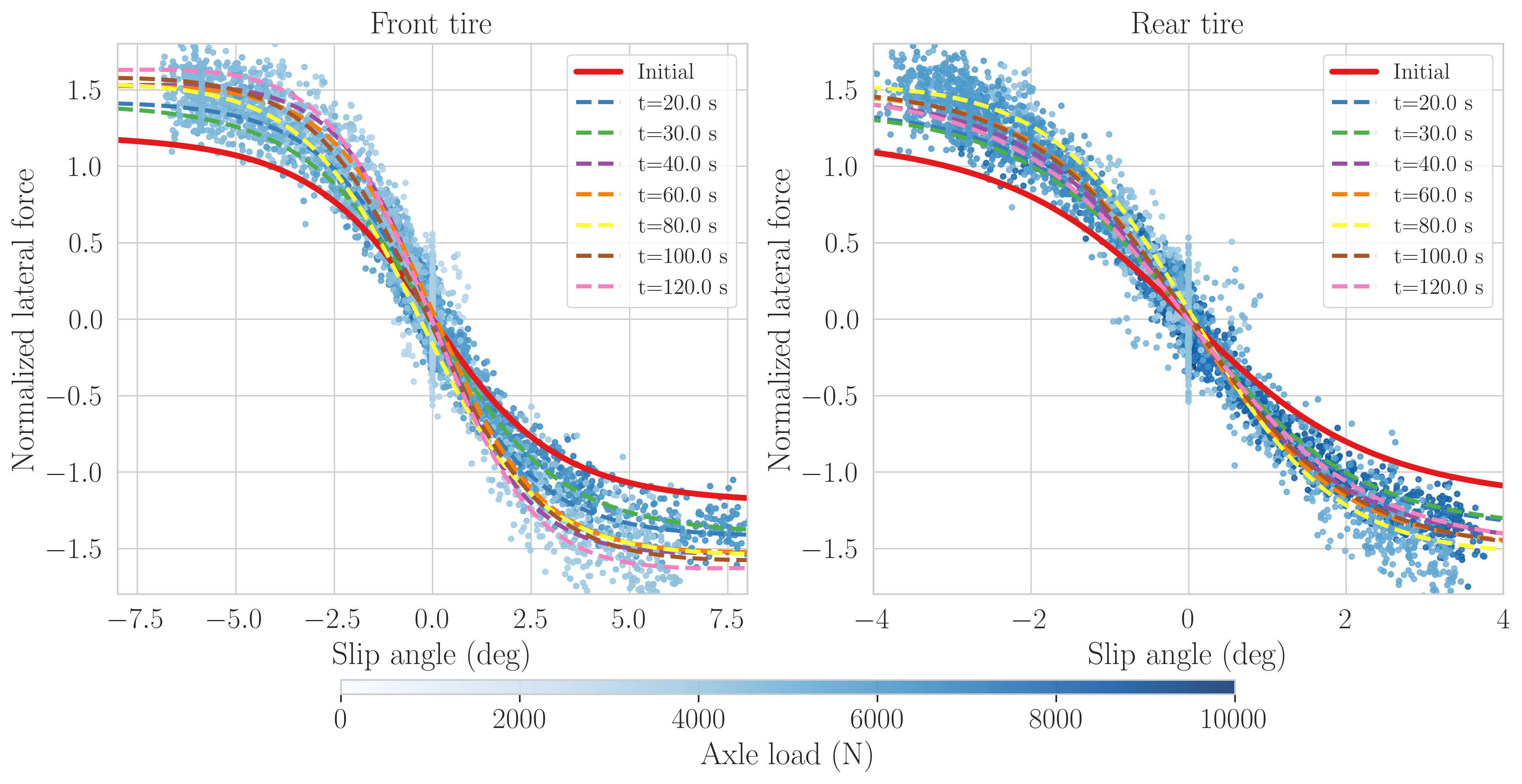}
    \caption{Front and rear tire Pacejka model fitted over time. The sequence of model parameters captures the changing tire behavior and lateral dynamics throughout the lap.}
    \label{fig:sim_pacejka}
\end{figure}

\begin{figure}[tb]
    \centering
    \includegraphics[width=\linewidth]{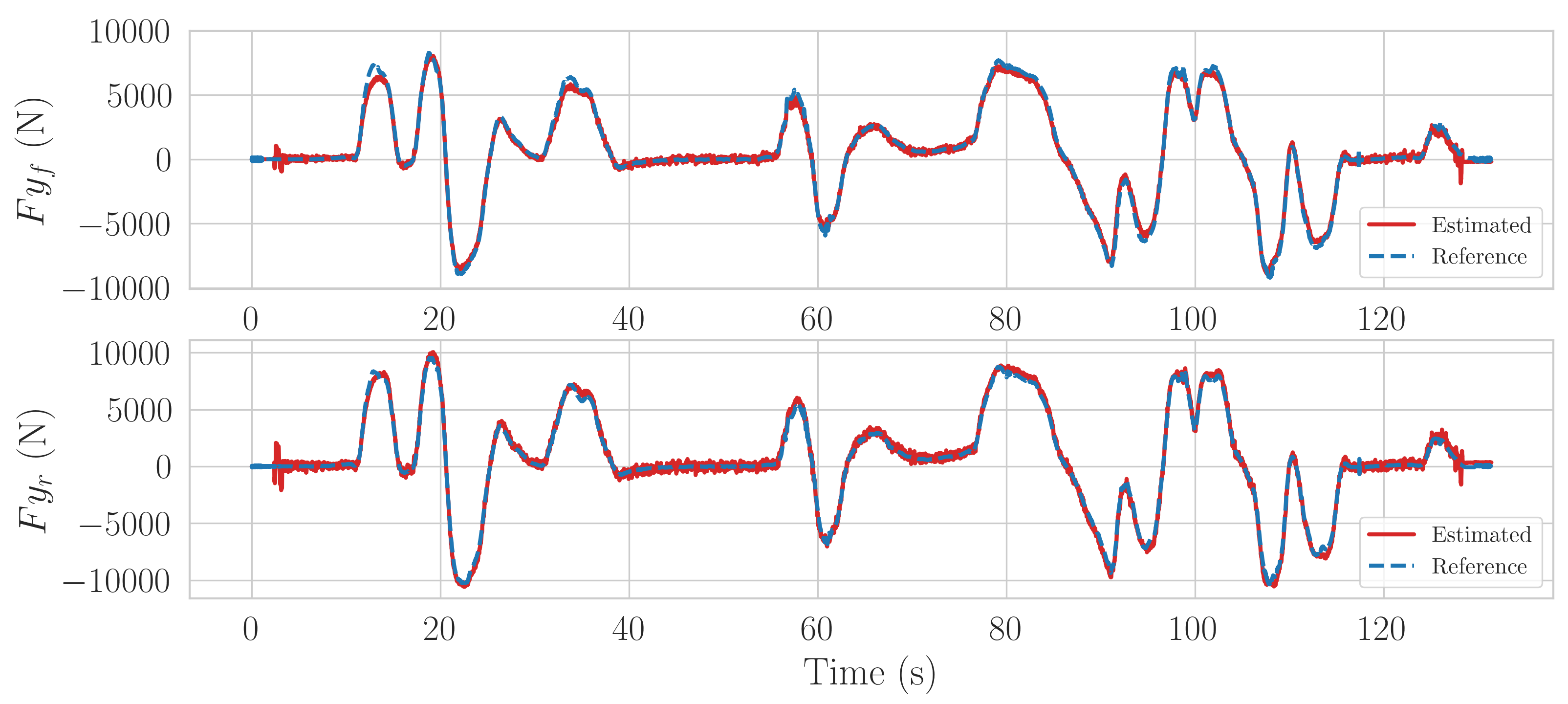}
    \caption{\nic{Estimated lateral forces during the hot-lap. RAGE tracks lateral forces accurately, with minor errors due to neglecting the limited-slip differential and combined slip effects.}}
    \label{fig:sim_lateral}
\end{figure}

For the final evaluation, a simulation of one hot-lap was performed on the Yas Marina Formula 1 Circuit. The initial Pacejka parameters for both the front and rear tires were randomized to assess RAGE’s performance in scenarios with unknown tire characteristics. This test aims to evaluate how quickly and accurately RAGE can adapt and identify valid parameters that describe the lateral dynamics of the vehicle.
Figure \ref{fig:sim_pacejka} illustrates the temporal evolution of the Pacejka tire model parameters throughout the simulation lap. The blue dots denote the normalized lateral force measurements acquired from the onboard IMU, capturing the tire's lateral behavior under various dynamic conditions. At the beginning of the lap, the model parameters are initialized randomly and consequently offer a poor fit to the actual tire responses. However, by approximately $t \sim 20$ seconds, shortly after the vehicle has completed the first two corners, RAGE has already converged to a significantly more accurate parameterization. From that point onward, RAGE updates the model parameters to capture the variation in tire behavior due to the changing cornering conditions experienced by the vehicle. The figure clearly illustrates that the sequence of identified parameterizations is able to accurately represent the wide range of lateral dynamics encountered throughout the lap.
Using the updated Pacejka tire model and the estimated slip angles, RAGE can reconstruct the lateral forces acting on each tire. As shown in Figure~\ref{fig:sim_lateral}, RAGE accurately tracks the lateral forces. However some estimation errors are observed. In particular, the front lateral forces are occasionally overestimated, while the rear lateral forces are underestimated. The observed mismatch is primarily attributed to the omission of the limited-slip differential's effect on the vehicle’s lateral force distribution in Equations (\ref{eq:fyf}) and (\ref{eq:fyr}). A minor mismatch also arises from neglecting longitudinal effects on lateral dynamics, which, although smaller, can still be significant when the tires experience combined slip conditions.
Nevertheless, despite these simplifications, the estimation framework demonstrates satisfactory performance. The root mean square error (RMSE) between the estimated lateral forces and the simulation reference remains below 500 N for both the front and rear axles, indicating that the model still captures the dominant dynamics with acceptable accuracy for most use cases.

\section{Experimental Results} \label{sec:results}
This section presents the experimental results obtained at the Yas Marina F1 Circuit during the A2RL event. In this context, the Kistler optical sensor is used exclusively as a reference to validate the estimated vehicle velocity, particularly its lateral component, as it is the only sensor capable of directly measuring this quantity. To reduce noise, the optical sensor velocity data has been processed using a first-order Butterworth low-pass filter with a cutoff frequency of 10 Hz.

At present, no publicly available datasets offer a combination of high-speed scenarios with RADAR measurements, making direct comparisons with other state-of-the-art methods on standardized benchmarks difficult and less meaningful. Consequently, the results presented here are more anecdotal in nature. Nonetheless, the primary contribution of the proposed approach lies not solely in its raw estimation accuracy, but in its ability to adapt in real-time to changing track conditions and tire dynamics, highlighting its robustness and practical value for autonomous racing applications.

\subsection{Case study}
\begin{figure}[h!tb]
    \centering
    \includegraphics[width=\linewidth]{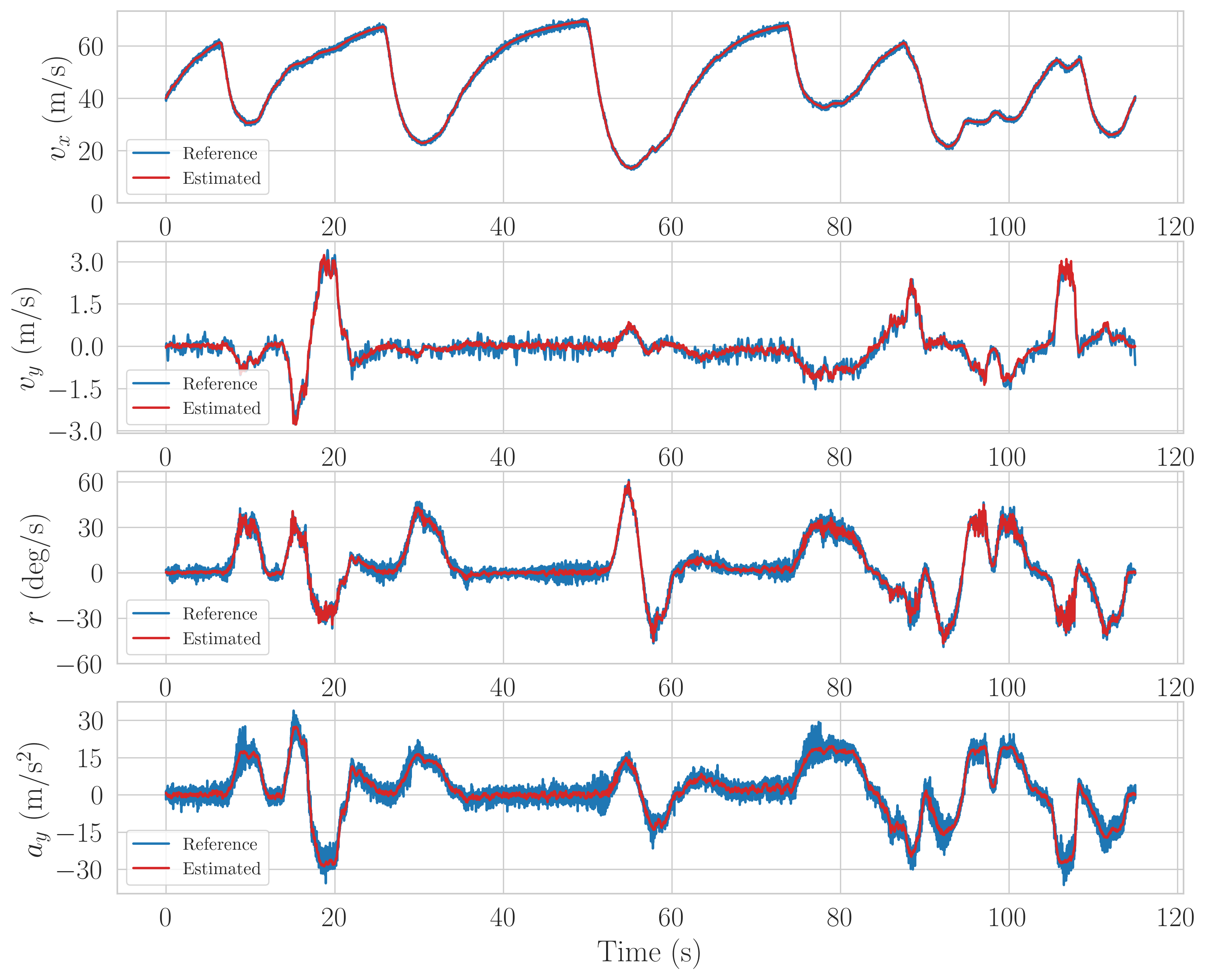}
    \caption{RAGE outputs compared with reference data obtained from the optical sensor, covering a complete lap of the Yas Marina F1 track. The lateral acceleration estimate is computed combining the front and rear lateral tire forces obtained from the Pacejka model with the online-estimated parameters.}
    \label{fig:real_fast_lap}
\end{figure}

In Figure \ref{fig:real_fast_lap} a complete lap of the Yas Marina F1 circuit is shown, during which the vehicle reaches speeds up to 70 m/s and lateral accelerations of up to 28 m/s$^2$. The outputs of the RAGE estimator exhibit strong agreement with the reference measurements, demonstrating its capability to accurately reconstruct both longitudinal and lateral velocities from RADAR Doppler data, even under high-speed maneuvers.

The lateral acceleration estimate $a_y$ is derived combining the estimated front and rear lateral tire forces, $F_{yf}$ and $F_{yr}$, respectively. This result confirms that the online-estimated Pacejka parameters effectively capture the lateral dynamics while attenuating measurements noise.

\begin{figure*}[tb]
    \centering
    \includegraphics[width=\textwidth]{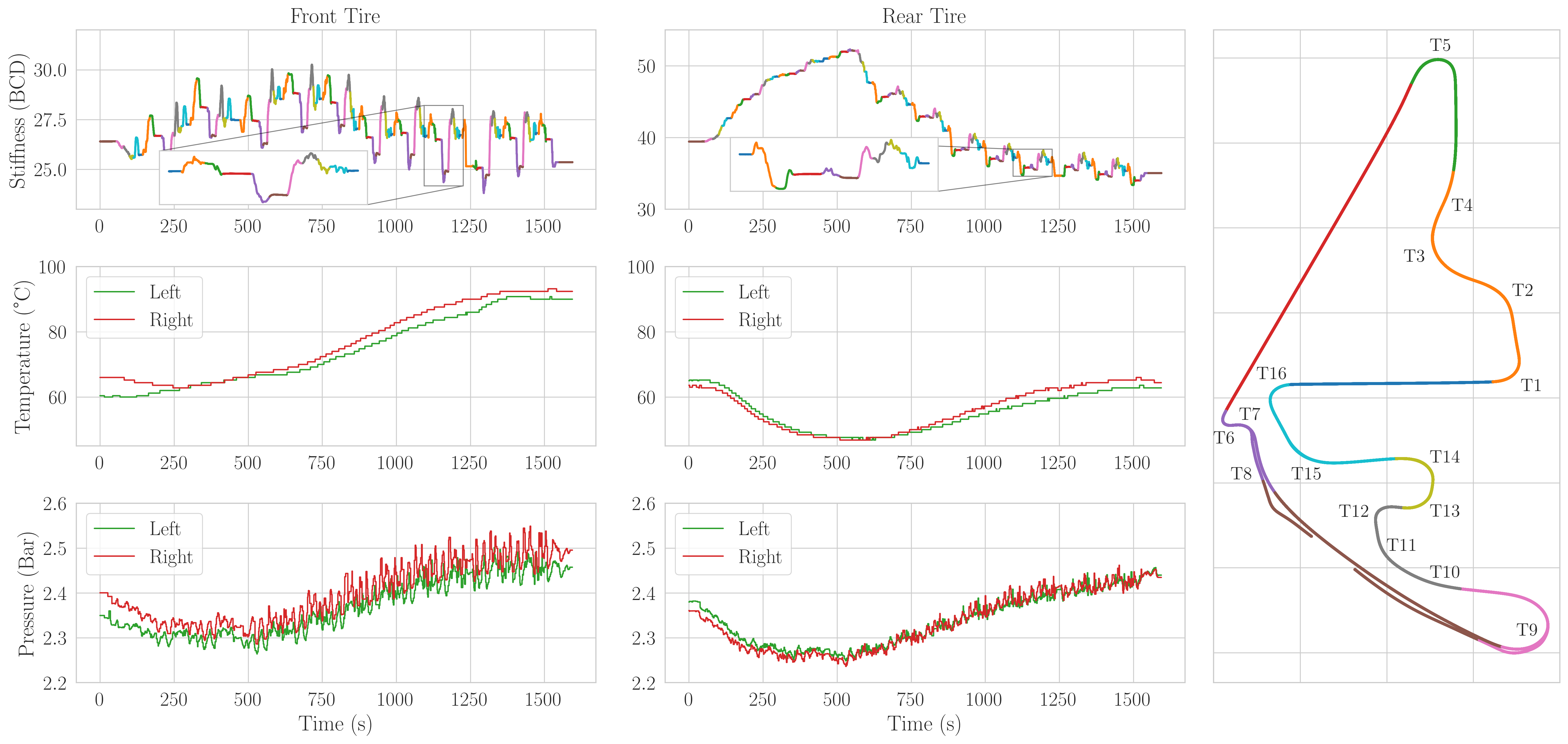}
    \caption{The estimated tire cornering stiffness shows a correlation with tire temperature and pressure, and exhibits periodic variations aligned with the circuit layout. The color-coded stiffness values correspond to specific segments of the track layout displayed on the right.}
    \label{fig:real_stiffness}
\end{figure*}

We further investigate the relationship between the estimated stiffness of the tire corners and the temperature and pressure of the tire, as measured by the Tire Pressure Monitoring System (TPMS) embedded in each tire. The cornering stiffness is computed from the estimated Pacejka parameters as $BCD$, with initial parameters obtained by an offline calibration using data from previous test sessions. These initial values characterize the tire-road interaction curve across a range of temperatures.
Figure \ref{fig:real_stiffness} presents data from a 26-minute evening session. At the beginning of the session, all tires were preheated to approximately 65°C using tire warmers, while the ambient track temperature was around 38.4°C. The session started immediately after mounting the heated tires on the EAV-24 vehicle. As shown in the figure, the front tires experience a temperature increase due to sustained braking and steering demands. In contrast, the temperature of the rear tires initially decreases, reaching a minimum of 45°C, driven by lower ambient temperatures and a front-biased brake distribution. After approximately $t\sim750$ seconds, the temperatures of the rear tires begin to rise again, eventually returning to 65°C approximately. Tire pressure closely follow the corresponding temperature profiles.
The cornering stiffness estimated by RAGE shown at the top of Figure \ref{fig:real_stiffness} tracks the variations in temperature and pressure, particularly for the rear tires. As expected, a clear inverse relationship is observed: as temperature and pressure decrease, stiffness increases, and vice versa \nic{\cite{farroni_temp,Teodosio_temp}}. In addition, the estimated stiffness values for the front and rear tires display periodic fluctuations that are spatially correlated with the layout of the track. These variations are visualized using color-coded stiffness values mapped to specific segments of the track layout, shown on the right. A zoomed-in temporal view of one full lap further illustrates that cornering stiffness remains approximately constant on straight sections but exhibits pronounced changes within corners. In particular, in high-speed turns (e.g. T1, T2, T3, T4 and T9), the variations are more significant due to the increased sensitivity of tire parameters to dynamic vertical loads.
This analysis highlights the ability of RAGE to dynamically adapt tire parameters in response to evolving vehicle, tire, and environmental conditions, allowing robust estimation in diverse operating scenarios.

\subsection{Computational Cost}
Finally, we analyze the computational cost of the complete estimation pipeline. All timing measurements were conducted on the EAV-24 computing platform, equipped with an AMD® EPYC 7003 "Milan" series processor and 128 GB of RAM. The implementation runs entirely on a single CPU core. The MHE relies on the CERES solver \cite{CERES}. To guarantee low-latency execution, the solver was configured with a maximum runtime of 8 ms and a limit of 3 iterations. In the evaluated case study, the solver achieved an average execution time of 2.287 ms, while all other components remained well below 0.5 ms. As expected, the optimization step dominates the computational load, nevertheless, the results confirm that the estimator can reliably operate at 100 Hz with a substantial performance margin.
\section{Conclusions and Future Work} \label{sec:conclusion}
In this paper, we introduced RAGE, a tightly coupled estimator that simultaneously estimates the vehicle’s velocity vector, tire slip angles, and lateral tire forces at both the front and rear axles. The proposed framework addresses key challenges associated with using RADAR sensors in high-speed environments, including measurement delay compensation and Doppler velocity de-aliasing.

The framework was initially validated in a high-fidelity simulation environment, where accurate ground truth data for each estimated parameter was available. We then assessed the system’s adaptability in real-world scenarios, where vehicle-tire-road friction varied dynamically with changes in tire temperature, pressure, and track layout.

Future research will focus on integrating a higher-fidelity vehicle dynamics model capable of estimating individual wheel slip angles and lateral forces. In parallel, a more sophisticated tire model will be introduced to estimate the underlying Pacejka micro-parameters, enabling the characterization of tire behavior under varying vertical loads. In addition, the framework will be extended to incorporate longitudinal dynamics, allowing modeling of combined slip conditions and their effects on tire force generation.
\section{Acknowledgements}
The authors would like to thank Micaela Verucchi and Ayoub Raji for theirs valuable comments which improved this manuscript.

\bibliographystyle{ieeetr} 
\bibliography{sections/refs} 

@INPROCEEDINGS{RAJI_PLANNING,
  author={Raji \textit{et al.}, Ayoub },
  booktitle={2022 IEEE 25th International Conference on Intelligent Transportation Systems (ITSC)}, 
  title={Motion Planning and Control for Multi Vehicle Autonomous Racing at High Speeds}, 
  year={2022},
  volume={},
  number={},
  pages={2775-2782},
  doi={10.1109/ITSC55140.2022.9922239}}

@article { DEALIAS,
      author = "Larry  Hennington",
      title = "Reducing the Effects of Doppler Radar Ambiguities",
      journal = "Journal of Applied Meteorology and Climatology",
      year = "1981",
      publisher = "American Meteorological Society",
      address = "Boston MA, USA",
      volume = "20",
      number = "12",
      doi = "10.1175/1520-0450(1981)020<1543:RTEODR>2.0.CO;2",
      url = "https://journals.ametsoc.org/view/journals/apme/20/12/1520-0450_1981_020_1543_rteodr_2_0_co_2.xml"
}

@article{CAMERA_SLIP_CORRELATION,
title = {Real-time side-slip angle measurements using digital image correlation},
journal = {Journal of Terramechanics},
volume = {81},
pages = {35-42},
year = {2019},
issn = {0022-4898},
doi = {https://doi.org/10.1016/j.jterra.2018.08.001},
url = {https://www.sciencedirect.com/science/article/pii/S002248981730263X},
author = {Devin K. Johnson and Theunis R. Botha and P. Schalk Els},
}

@INPROCEEDINGS{CAMERA_SLIP_RC_SCALED,
  author={Kuyt, Conno and Como, Matteo},
  booktitle={2018 IEEE Conference on Control Technology and Applications (CCTA)}, 
  title={Mixed Kinematics and Camera Based Vehicle Dynamic Sideslip Estimation for an RC Scaled Model}, 
  year={2018},
  volume={},
  number={},
  doi={10.1109/CCTA.2018.8511487}}

@InProceedings{CAMERA_SLIP_DOWNFACE,
author="Serena, L.
and Bruschetta, M.
and de Castro, R.
and Lenzo, B.",
editor="Quaglia, Giuseppe
and Boschetti, Giovanni
and Carbone, Giuseppe",
title="Estimating Sideslip Angle Using a Downward-Facing Camera",
booktitle="Advances in Italian Mechanism Science",
year="2024",
publisher="Springer Nature Switzerland",
address="Cham",
pages="290--297",
}

@article{CAMERA_SLIP_DOWNFACE2,
title = {Real time camera-based sideslip angle estimation: design and experiments},
journal = {IFAC-PapersOnLine},
volume = {58},
number = {28},
pages = {750-755},
year = {2024},
issn = {2405-8963},
doi = {https://doi.org/10.1016/j.ifacol.2025.01.056},
url = {https://www.sciencedirect.com/science/article/pii/S2405896325000564},
author = {Leonardo Serena and Mattia Bruschetta and Giovanni Righetti and Ricardo {de Castro} and Basilio Lenzo},
}

@INPROCEEDINGS{CAMERA_SLIP_DOWNFACE3,
  author={Serena, Leonardo and Lenzo, Basilio and Bruschetta, Mattia and De Castro, Ricardo},
  booktitle={2023 IEEE International Workshop on Metrology for Automotive}, 
  title={Computer vision approaches for vehicle sideslip angle estimation}, 
  year={2023},
  volume={},
  number={},
  doi={10.1109/MetroAutomotive57488.2023.10219124}}

@inproceedings{LOP_UKF,
   title={Guess the Drift with LOP-UKF: LiDAR Odometry and Pacejka Model for Real-Time Racecar Sideslip Estimation},
   url={http://dx.doi.org/10.1109/IV55156.2024.10588524},
   DOI={10.1109/iv55156.2024.10588524},
   booktitle={2024 IEEE Intelligent Vehicles Symposium (IV)},
   publisher={IEEE},
   author={Toschi \textit{et al.}, Alessandro},
   year={2024},
   month=jun, pages={885–891} }

@article{GNSS_SLIP,
title = {Vehicle sideslip angle estimation by fusing inertial measurement unit and global navigation satellite system with heading alignment},
journal = {Mechanical Systems and Signal Processing},
volume = {150},
pages = {107290},
year = {2021},
issn = {0888-3270},
doi = {https://doi.org/10.1016/j.ymssp.2020.107290},
url = {https://www.sciencedirect.com/science/article/pii/S0888327020306762},
author = {Xin Xia and Lu Xiong and Yishi Lu and Letian Gao and Zhuoping Yu},
}

@article{GNSS_DOUBLE_ANTENNA_SLIP,
title = {Vehicle sideslip angle measurement based on sensor data fusion using an integrated ANFIS and an Unscented Kalman Filter algorithm},
journal = {Mechanical Systems and Signal Processing},
volume = {72-73},
pages = {832-845},
year = {2016},
issn = {0888-3270},
doi = {https://doi.org/10.1016/j.ymssp.2015.11.003},
url = {https://www.sciencedirect.com/science/article/pii/S0888327015004999},
author = {B.L. Boada and M.J.L. Boada and V. Diaz},
}

@INPROCEEDINGS{GNSS_SLIP2,
  author={Binh Minh Nguyen and Yafei Wang and Fujimoto, Hiroshi and Hori, Yoichi},
  booktitle={2012 IEEE Vehicle Power and Propulsion Conference}, 
  title={Sideslip angle estimation using gps and disturbance accommodating multi-rate Kalman filter for electric vehicle stability control}, 
  year={2012},
  volume={},
  number={},
  pages={1323-1328},
  doi={10.1109/VPPC.2012.6422741}}

@article{GNSS_SLIP3,
author = {Takikawa, Kanamu and Atsumi, Yoshiki and Takanose, Aoki and Meguro, Junichi},
year = {2021},
month = {02},
pages = {},
title = {Vehicular trajectory estimation utilizing slip angle based on GNSS Doppler/IMU},
volume = {8},
journal = {ROBOMECH Journal},
doi = {10.1186/s40648-021-00195-4}
}

@article{GNSS_SLIP4,
 author    = {Wenkang Wan and Kai Sheng and Qing Cai and Lei Ao and Nan Ouyang and Haoxuan Feng and Dejene M. Sime},
  title     = {Enhanced tire road friction coefficient estimation through interacting multiple model design based on tire force observation},
  journal   = {Nonlinear Dynamics},
  volume    = {113},
  number    = {13},
  pages     = {15925--15941},
  year      = {2025},
  month     = jul,
  doi       = {10.1007/s11071-025-10867-0},
  url       = {https://doi.org/10.1007/s11071-025-10867-0},
  issn      = {1573-269X}
}

@Article{DYNAMIC_KINEMATIC_MODEL,
AUTHOR = {Righetti, Giovanni and Lenzo, Basilio},
TITLE = {A Combined Dynamic–Kinematic Extended Kalman Filter for Estimating Vehicle Sideslip Angle},
JOURNAL = {Applied Sciences},
VOLUME = {15},
YEAR = {2025},
NUMBER = {3},
ARTICLE-NUMBER = {1365},
URL = {https://www.mdpi.com/2076-3417/15/3/1365},
ISSN = {2076-3417},
DOI = {10.3390/app15031365}
}

@article{KINEMATIC_MODEL,
title = {Vehicle sideslip estimation: A kinematic based approach},
journal = {Control Engineering Practice},
volume = {67},
pages = {1-12},
year = {2017},
issn = {0967-0661},
doi = {https://doi.org/10.1016/j.conengprac.2017.06.013},
url = {https://www.sciencedirect.com/science/article/pii/S0967066117301491},
author = {Donald Selmanaj and Matteo Corno and Giulio Panzani and Sergio M. Savaresi},
}

@article{MODEL_BASED,
doi = {10.1088/1742-6596/2090/1/012156},
url = {https://dx.doi.org/10.1088/1742-6596/2090/1/012156},
year = {2021},
month = {nov},
publisher = {IOP Publishing},
volume = {2090},
number = {1},
pages = {012156},
author = {D’Inverno, M and Arricale, V M and Zanardi, A and Frazzoli, E and Sakhnevych, A and Timpone, F},
title = {A benchmark study on the model-based estimation of the go-kart side-slip angle},
journal = {Journal of Physics: Conference Series},
}

@article{DYNAMIC_KINEMATIC_MODEL2,
author = {Villano, Elvis and Lenzo, Basilio and Sakhnevych, Aleksandr},
year = {2021},
month = {09},
pages = {},
title = {Cross-combined UKF for vehicle sideslip angle estimation with a modified {D}ugoff tire model: design and experimental results},
volume = {56},
journal = {Meccanica},
doi = {10.1007/s11012-021-01403-6}
}

@article{TRICK2,
author = {Napolitano Dell’Annunziata, Guido and Farroni, Flavio and Timpone, Francesco and Lenzo, Basilio},
year = {2025},
month = {04},
pages = {},
title = {T.R.I.C.K. 2.0: Enhanced Vehicle Dynamics Analysis and Estimation Harnessing Advanced Vehicle Sensors},
volume = {9},
journal = {SAE International Journal of Vehicle Dynamics, Stability, and NVH},
doi = {10.4271/10-09-03-0025}
}

@article{PACEJKA,
  title={THE MAGIC FORMULA TYRE MODEL},
  author={Hans B. Pacejka and Egbert Bakker},
  journal={Vehicle System Dynamics},
  year={1991},
  volume={21},
  pages={1-18},
  url={https://api.semanticscholar.org/CorpusID:108456393}
}

@misc{CREVE,
      title={CREVE: An Acceleration-based Constraint Approach for Robust Radar Ego-Velocity Estimation}, 
      author={Hoang Viet Do and Bo Sung Ko and Yong Hun Kim and Jin Woo Song},
      year={2025},
      eprint={2409.16847},
      archivePrefix={arXiv},
      primaryClass={cs.RO},
      url={https://arxiv.org/abs/2409.16847}, 
}

@misc{RAVE,
      title={RAVE: A Framework for Radar Ego-Velocity Estimation}, 
      author={Vlaho-Josip Štironja and Luka Petrović and Juraj Peršić and Ivan Marković and Ivan Petrović},
      year={2024},
      eprint={2406.18850},
      archivePrefix={arXiv},
      primaryClass={cs.RO},
      url={https://arxiv.org/abs/2406.18850}, 
}

@ARTICLE{LIRO,
  author={Li, Shengyu and Li, Xingxing and Chen, Shuolong and Zhou, Yuxuan and Wang, Shiwen},
  journal={IEEE Internet of Things Journal}, 
  title={Continuous-Time LiDAR/IMU/Radar Odometry for Accurate and Smooth State Estimation via Tightly Coupled Integration}, 
  year={2024},
  doi={10.1109/JIOT.2024.3450654}}

@misc{GARLIO,
      title={GaRLIO: Gravity enhanced Radar-LiDAR-Inertial Odometry}, 
      author={Chiyun Noh and Wooseong Yang and Minwoo Jung and Sangwoo Jung and Ayoung Kim},
      year={2025},
      eprint={2502.07703},
      archivePrefix={arXiv},
      primaryClass={cs.RO},
      url={https://arxiv.org/abs/2502.07703}, 
}

@inproceedings{ROAMER,
  author={Nissov, Morten and Khattak, Shehryar and Edlund, Jeffrey A. and Padgett, Curtis and Alexis, Kostas and Spieler, Patrick},

  booktitle={2024 IEEE Aerospace Conference}, 

  title={ROAMER: Robust Offroad Autonomy using Multimodal State Estimation with Radar Velocity Integration}, 

  year={2024},

  volume={},

  number={},

  pages={1-10},

  doi={10.1109/AERO58975.2024.10521170}}

@article{RVF,
 author={Nissov, Morten and Edlund, Jeffrey A. and Spieler, Patrick and Padgett, Curtis and Alexis, Kostas and Khattak, Shehryar},

  journal={IEEE Robotics and Automation Letters}, 

  title={Robust High-Speed State Estimation for Off-Road Navigation Using Radar Velocity Factors}, 

  year={2024},

  volume={9},

  number={12},

  pages={11146-11153},

  doi={10.1109/LRA.2024.3486189}}

@inproceedings{RADAR_SIDE_SLIP_FILTER,
author = {Diener, Luis and Kalkkuhl, Jens and Schirle, Thomas},
year = {2024},
month = {07},
pages = {10},
conference={Stuttgart International Symposium },
title = {Radar-Based Approach for Side-Slip Gradient Estimation},
doi = {10.4271/2024-01-2976}
}

@inproceedings{MADGWICK,
  title={An efficient orientation filter for inertial and inertial / magnetic sensor arrays},
  author={Sebastian O. H. Madgwick},
  year={2010},
  url={https://api.semanticscholar.org/CorpusID:2976407}
}

@ARTICLE{ZUPT,

  author={Foxlin, E.},

  journal={IEEE Computer Graphics and Applications}, 

  title={Pedestrian tracking with shoe-mounted inertial sensors}, 

  year={2005},

  volume={25},

  number={6},

  pages={38-46},

  doi={10.1109/MCG.2005.140}}

@INPROCEEDINGS{PTP,
  author={Watt, Steve T. and Achanta, Shankar and Abubakari, Hamza and Sagen, Eric and Korkmaz, Zafer and Ahmed, Husam},
  booktitle={2015 Saudi Arabia Smart Grid (SASG)}, 
  title={Understanding and applying precision time protocol}, 
  year={2015},
  volume={},
  number={},
  pages={1-7},
  doi={10.1109/SASG.2015.7449285}}

@article{RADAR_ALIAS_DESC,
author = {Jr, Robert},
year = {2001},
pages = {1674-1683},
title = {A Real-Time Four-Dimensional Doppler Dealiasing Scheme},
volume = {18},
journal = {Journal of Atmospheric and Oceanic Technology},
doi = {10.1175/1520-0426(2001)018<1674:ARTFDD>2.0.CO;2}
}

@article{NN1,
author={Srinivasan, Sirish and Sa, Inkyu and Zyner, Alex and Reijgwart, Victor and Valls, Miguel I. and Siegwart, Roland},
  journal={IEEE Robotics and Automation Letters}, 
  title={End-to-End Velocity Estimation for Autonomous Racing}, 
  year={2020},
  volume={5},
  number={4},
  pages={6869-6875},
  doi={10.1109/LRA.2020.3016929}}

@article{NN2,
doi = {10.1088/1742-6596/1888/1/012006},
url = {https://dx.doi.org/10.1088/1742-6596/1888/1/012006},
year = {2021},
month = {apr},
publisher = {IOP Publishing},
volume = {1888},
number = {1},
pages = {012006},
author = {Chindamo, Daniel and Gadola, Marco and Bonera, Emanuele and Magri, Paolo},
title = {Experimental Comparison of The Two Most Used Vehicle Sideslip Angle Estimation Methods for Model-Based Design Approach},
journal = {Journal of Physics: Conference Series}
}

@article{NN3,
  author = "Giuliacci, Tiziano Alberto and Ballesio, Stefano and Fainello, Marco and Mair, Ulrich and King, Julian",
  title = "Recurrent Neural Network Model for On-Board Estimation of the
                    Side-Slip Angle in a Four-Wheel Drive and Steering Vehicle",
  journal = "SAE International Journal of Passenger Vehicle Systems",
  volume = "17",
  number = "1",
  pages = "37--48",
  publisher = "SAE International",
  month = sep,
  year = 2023,
  doi = "https://doi.org/10.4271/15-17-01-0003",
  url = "https://doi.org/10.4271/15-17-01-0003",
}

@software{CERES,
  author = {Agarwal, Sameer and Mierle, Keir and The Ceres Solver Team},
  title = {{Ceres Solver}},
  license = {Apache-2.0},
  url = {https://github.com/ceres-solver/ceres-solver},
  version = {2.2},
  year = {2023},
  month = {10}
}

@ARTICLE{ER_AUTOPILOT,
  author={Raji \textit{et al.}, Ayoub},
  journal={IEEE Transactions on Field Robotics}, 
  title={er.autopilot 1.1: A Software Stack for Autonomous Racing on Oval and Road Course Tracks}, 
  year={2024},
  volume={1},
  number={},
  pages={332-359},
  doi={10.1109/TFR.2024.3501252}}

@ARTICLE{ISO3888-1,
  title        = {{ISO 3888-1:2018} Passenger cars — Test track for a severe lane-change manoeuvre — Part 1: Double Lane-Change},
  organization = {International Organization for Standardization},
  year         = {2018},
  address      = {Geneva, Switzerland},
}

@book{guiggiani2023science,
  title     = {The Science of Vehicle Dynamics: Handling, Braking, and Ride of Road and Race Cars},
  author    = {Guiggiani, Massimo},
  year      = {2004},
  publisher = {Springer International Publishing},
  isbn      = {9783031064630},
  url       = {https://books.google.it/books?id=g0450AEACAAJ}
}

@article{farroni_temp,
author = {Flavio Farroni and Michele Russo and Aleksandr Sakhnevych and Francesco Timpone},
title ={TRT EVO: Advances in real-time thermodynamic tire modeling for vehicle dynamics simulations},

journal = {Proceedings of the Institution of Mechanical Engineers, Part D: Journal of Automobile Engineering},
volume = {233},
number = {1},
pages = {121-135},
year = {2019},
doi = {10.1177/0954407018808992},

}

@article{Teodosio_temp,
  author  = {Luigi Teodosio and Giuseppe Alferi and Andrea Genovese and Flavio Farroni and Benedetto Mele and Francesco Timpone and Aleksandr Sakhnevych},
  title   = {A numerical methodology for thermo-fluid dynamic modelling of tyre inner chamber: towards real time applications},
  journal = {Meccanica},
  year    = {2021},
  volume  = {56},
  number  = {3},
  pages   = {549--567},
  doi     = {10.1007/s11012-021-01310-w},
  url     = {https://doi.org/10.1007/s11012-021-01310-w}
}
\vspace{-1cm}
\begin{IEEEbiography}[{\includegraphics[width=1in,height=1.2in,keepaspectratio,trim={500px 1300px 500px 300px},clip]{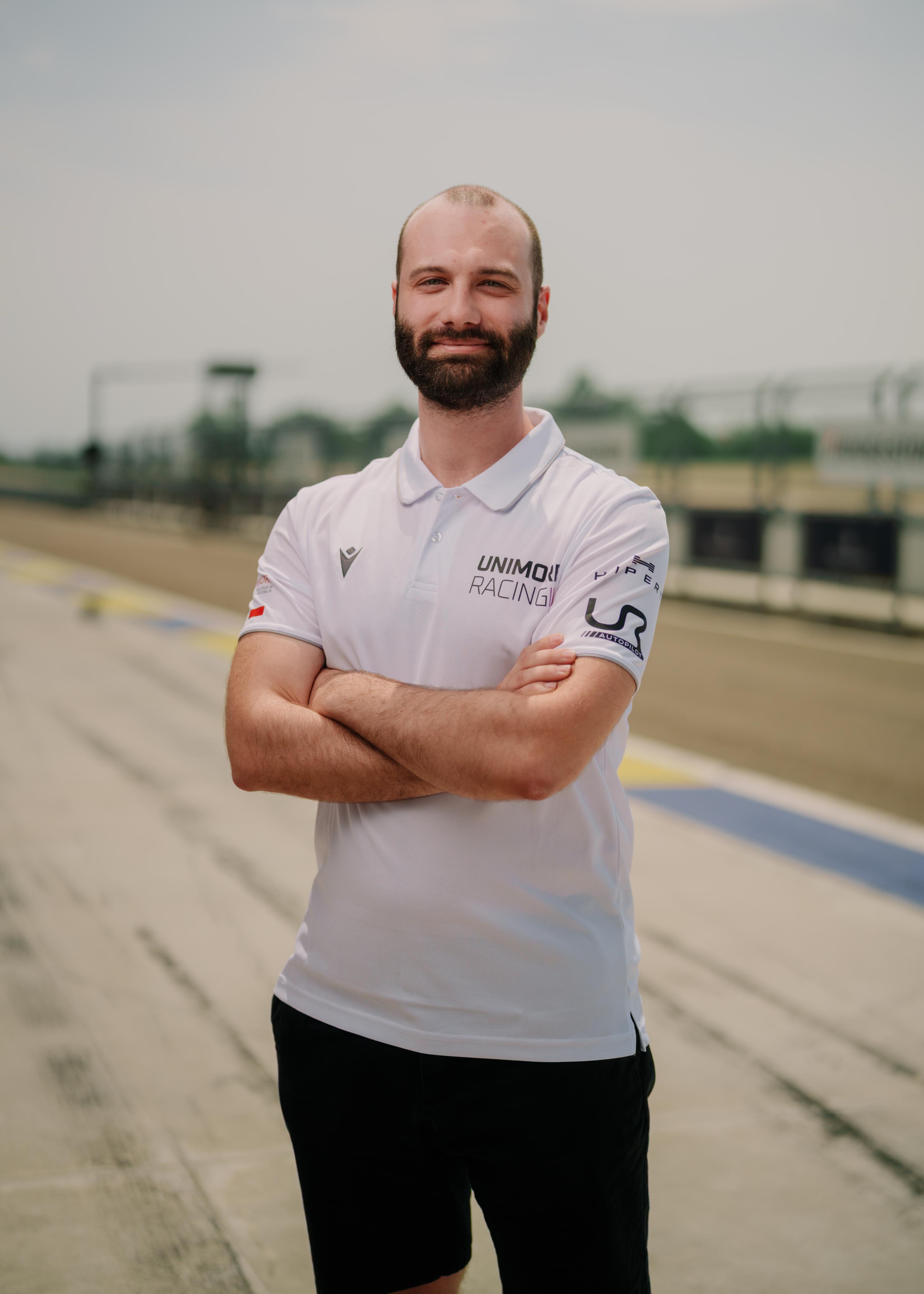}}]{Davide Malvezzi } received his M.S. degree in Computer Science from the University of Modena and Reggio Emilia, Italy, in 2020. He is currently working toward the Ph.D. degree with the Department of Physics, Mathematics and Computer Science, at the same institution. His doctoral research focuses on the development of perception and state estimation algorithms tailored for autonomous racing vehicles.
\end{IEEEbiography}
\vspace{-1.1cm}
\begin{IEEEbiography}[{\includegraphics[width=1in,height=1.2in,keepaspectratio,trim={410px 1400px 410px 300px},clip]{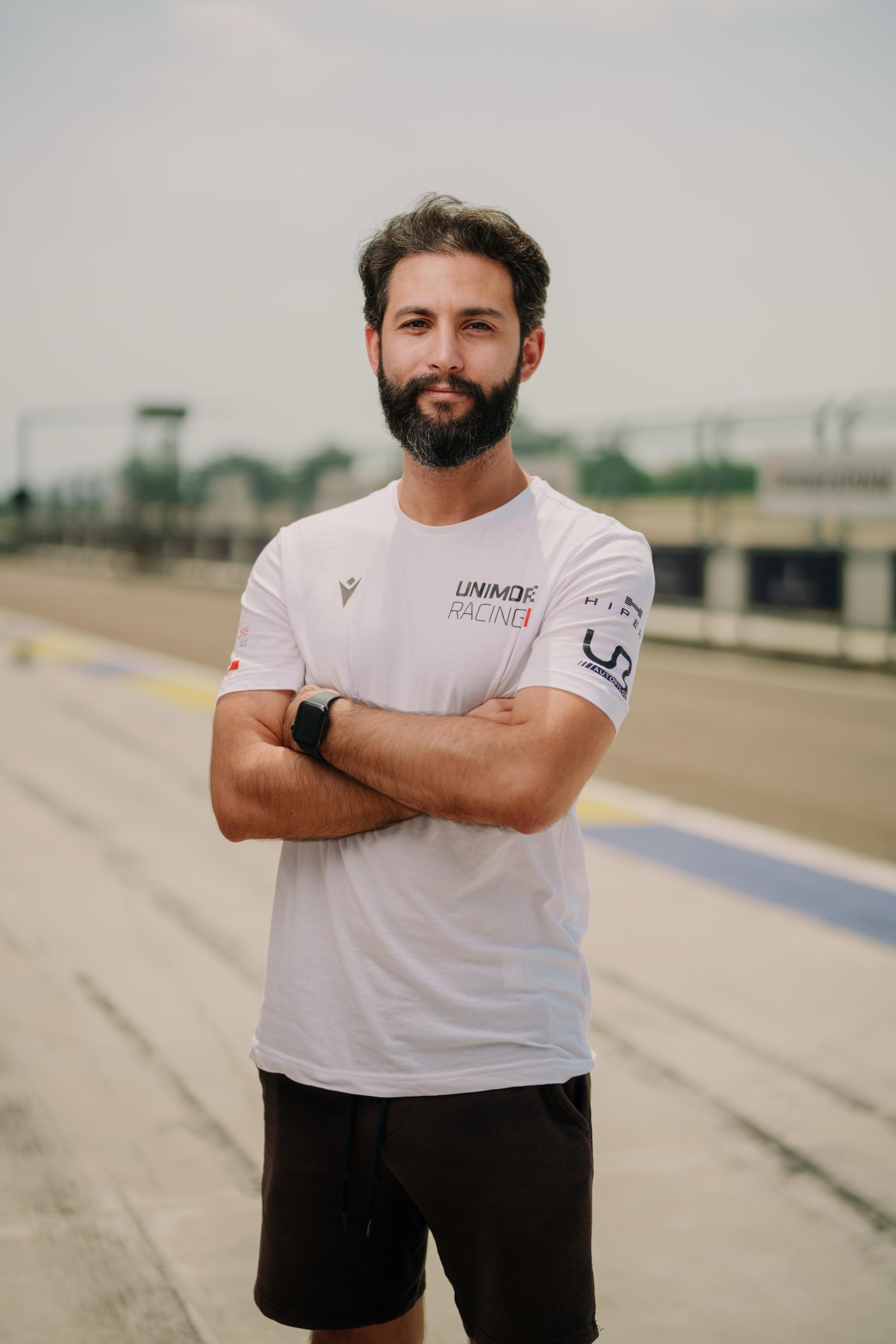}}]{Nicola Musiu} received his M.S. degree in Vehicle Engineering from the University of Modena and Reggio Emilia, Italy, in 2022. He is currently working toward the Ph.D. degree with the Department of Physics, Mathematics, and Computer Science at the same institution. His research focuses on vehicle dynamics modeling, multi-body simulation and model predictive control for autonomous racing vehicles.\end{IEEEbiography}
\vspace{-1.1cm}
\begin{IEEEbiography}[{\includegraphics[width=1in,height=1.2in,keepaspectratio,trim={410px 1400px 410px 300px},clip]{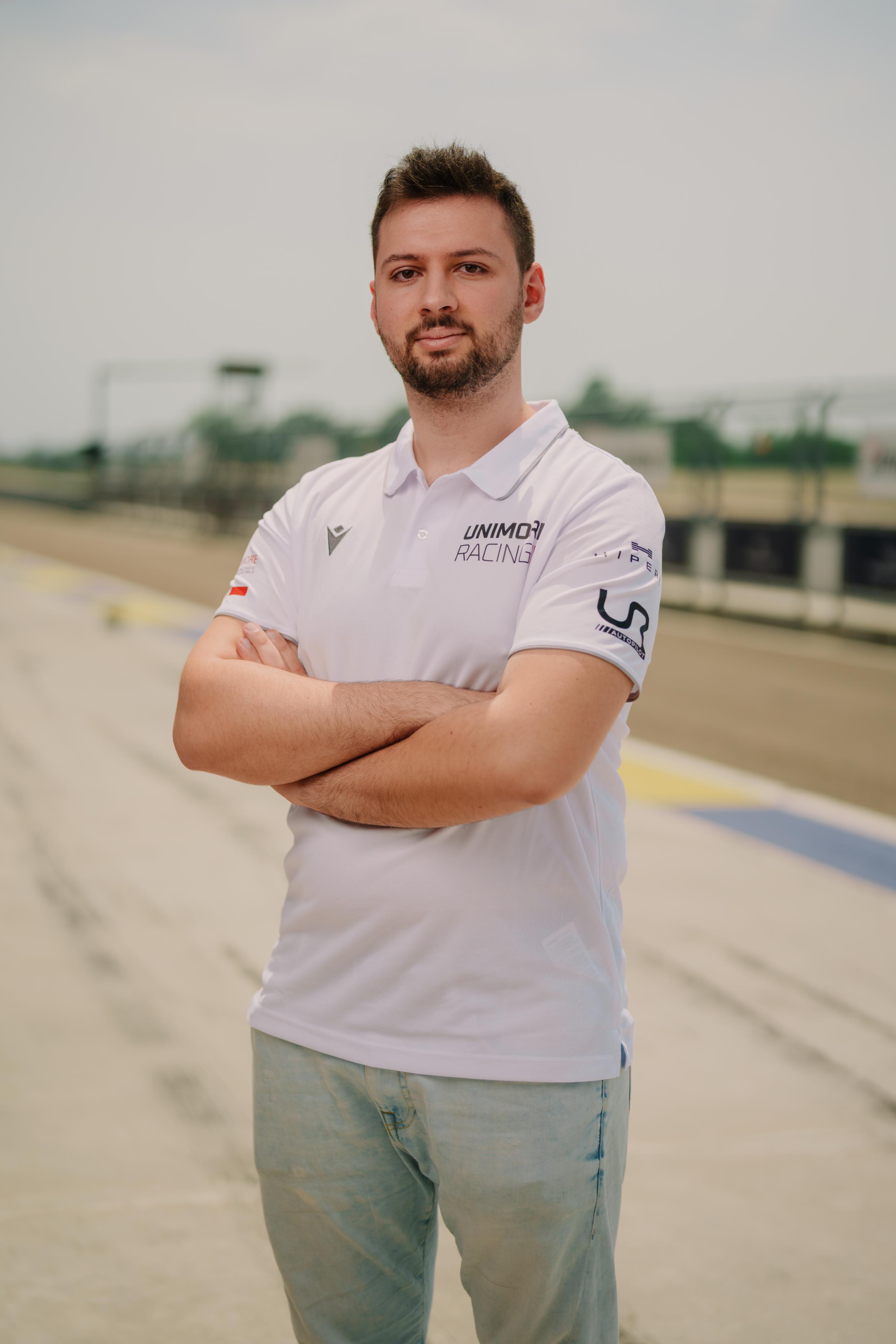}}]{Eugenio Mascaro} received his M.S. degree in Vehicle Engineering from the University of Modena and Reggio Emilia, Italy, in 2023. He is currently working as a Research Fellow with the Department of Physics, Mathematics, and Computer Science at the same institution. His research focuses on vehicle dynamics modeling and multi-body simulation for autonomous racing vehicles.\end{IEEEbiography}
\vspace{-1.1cm}
\begin{IEEEbiography}[{\includegraphics[width=1in,height=1.2in,keepaspectratio,trim={480px 1400px 550px 300px},clip]{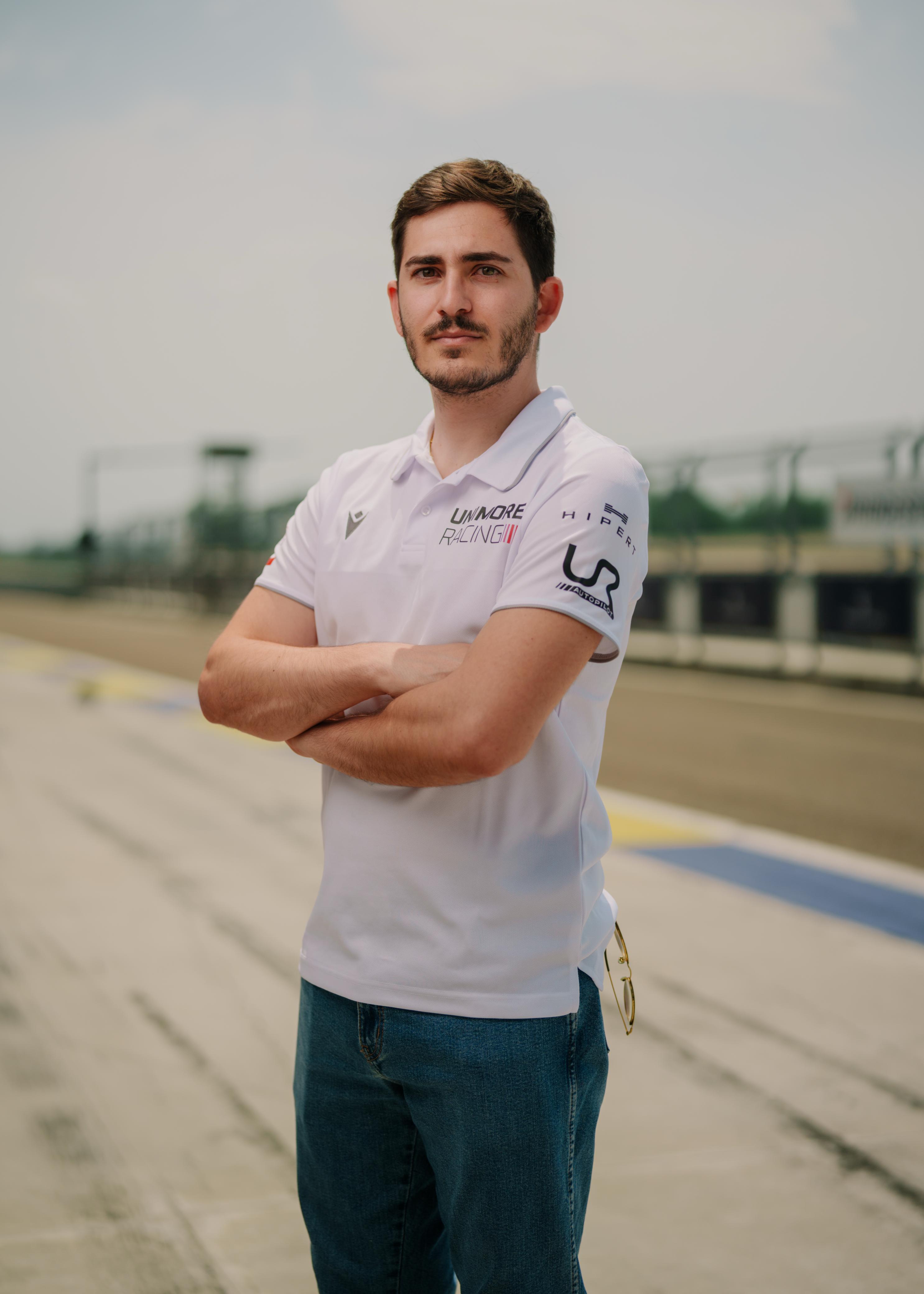}}]{Francesco Iacovacci}
received his M.S. degree in Vehicle Engineering from the University of Modena and Reggio Emilia, Italy, in 2022. He is currently working as a Research Fellow with the Department of Physics, Mathematics, and Computer Science at the same institution. His research focuses on model identification and performance analysis for autonomous racing vehicles.\end{IEEEbiography}
\vspace{-1.1cm}
\begin{IEEEbiography}[{\includegraphics[width=1in,height=1.2in,keepaspectratio,trim={400px 1400px 440px 300px},clip]{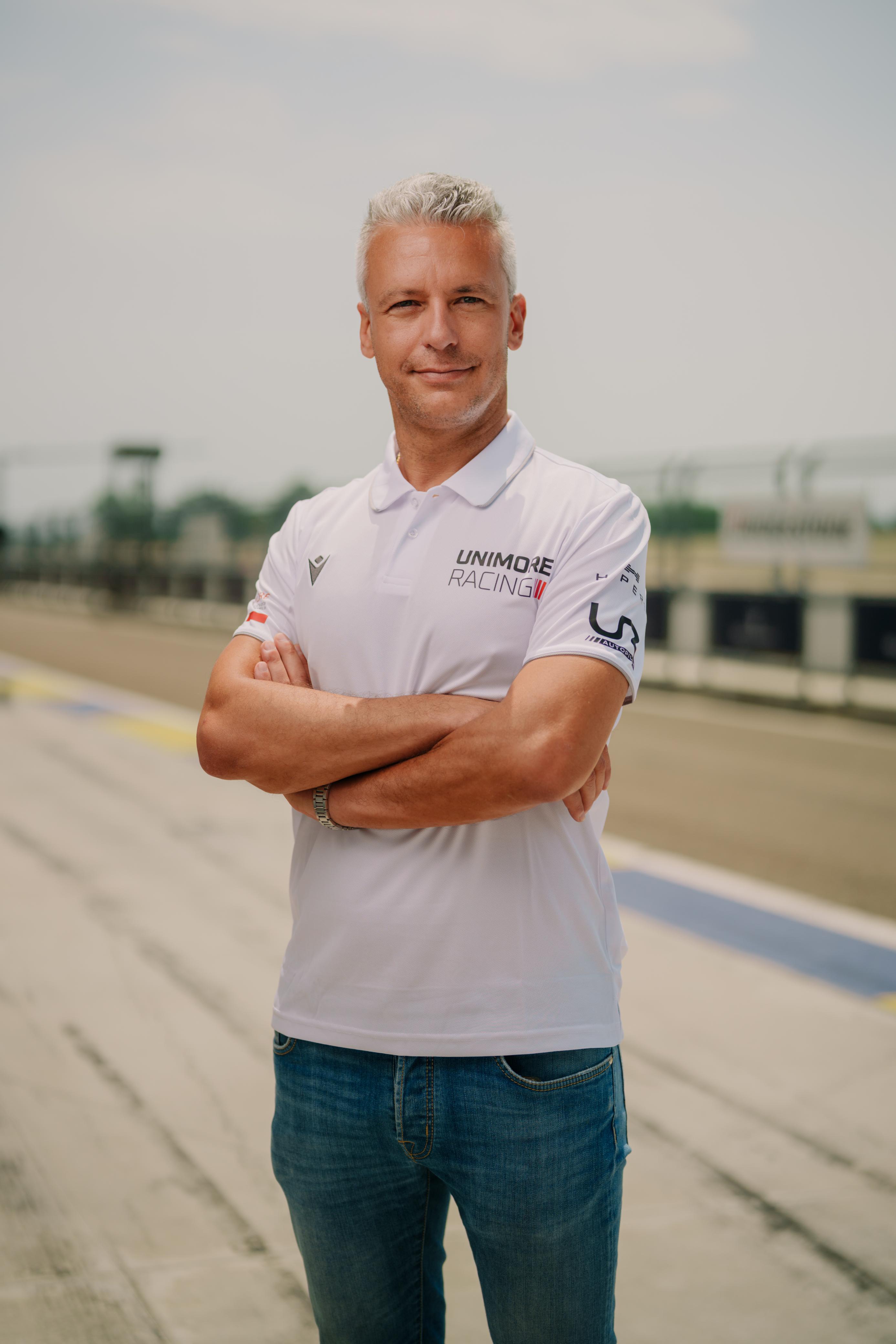}}]{Marko Bertogna} is Full Professor at University of Modena and Reggio Emilia, Italy, and leader of the HiPeRT Lab. His main research interests are in High-Performance Real-Time systems, especially based on multi- and many-core devices, Autonomous Driving and Industrial Automation. In 2008, he received a PhD in Computer Sciences from the Scuola Superiore Sant'Anna of Pisa, Italy. He has authored more than 100 papers, receiving multiple Best Paper Awards in first level international conferences. He coordinated multiple EU and industrial projects, securing more than 15 MEuro in funding for his research group. He is CEO and founder of the academic spin-off HiPeRT Srl.\end{IEEEbiography}

\end{document}